\def\year{2021}\relax
\documentclass[letterpaper]{article} 
\usepackage{aaai21}  
\usepackage{times}  
\usepackage{helvet} 
\usepackage{courier}  
\usepackage[hyphens]{url}  
\usepackage{graphicx} 
\usepackage{natbib}  
\usepackage{caption} 
\usepackage{amsmath}
\usepackage{array}
\usepackage{booktabs}
\usepackage{multicol}
\usepackage{enumitem}
\usepackage{subcaption}
\usepackage{xr}
\usepackage{xtab}

\newcolumntype{L}[1]{>{\raggedright\let\newline\\\arraybackslash\hspace{0pt}}m{#1}}
\newcolumntype{C}[1]{>{\centering\let\newline\\\arraybackslash\hspace{0pt}}m{#1}}
\newcolumntype{R}[1]{>{\raggedleft\let\newline\\\arraybackslash\hspace{0pt}}m{#1}}

\title{Dialog Policy Learning for Joint Clarification and Active Learning Queries}
\author{
Aishwarya Padmakumar, \textsuperscript{\rm 1}
Raymond J. Mooney \textsuperscript{\rm 2} \\
}
\affiliations{
\textsuperscript{\rm 1} Amazon Alexa AI~\footnote{This work was done as a part of the first author's PhD at the University of Texas at Austin} \\
\textsuperscript{\rm 2} University of Texas at Austin \\
padmakua@amazon.com, mooney@cs.utexas.edu
}

\begin{document}

\maketitle

\begin{abstract}
Intelligent systems need to be able to recover from mistakes, resolve uncertainty, and adapt to novel concepts not seen during training. 
Dialog interaction can enable this by the use of clarifications for correction and resolving uncertainty, and active learning queries to learn new concepts encountered during operation.
Prior work on dialog systems has either focused on exclusively learning how to perform clarification/ information seeking, or to perform active learning. 
In this work, we train a hierarchical dialog policy to jointly perform {\it both} clarification and active learning in the context of an interactive language-based image retrieval task motivated by an online shopping application, and demonstrate that jointly learning dialog policies for clarification and active learning is more effective than the use of static dialog policies for one or both of these functions.
\end{abstract}

\section{Introduction}

The ability to understand and communicate in natural language can improve the accessibility of systems such as robots, home devices and computers to non-expert users. 
Since language is often be ambiguous, it is desirable for such systems to engage in a dialog with the user to clarify their intentions and obtain missing information. 
We use \textit{clarification} to refer to any dialog act that enables the system to better understand an ongoing user request.
Common clarification questions obtain or clarify the value of a slot or argument that is part of a goal the user is trying to communicate.

A particular application may also contain domain-specific vocabulary or concepts that were not encountered during training.
For example, a system in a shopping domain may need to be updated with the introduction of new clothing styles.
Hence, it is desirable for a system to adapt to the operating environment using information from user interactions.
We use the term \textit{active learning} to refer to dialog acts used to obtain such knowledge with the primary purpose of improving the underlying language understanding model and thereby improving performance on future interactions.

Prior work on dialog and user interaction typically focuses either exclusively on clarification~\cite{young:procieee13,padmakumar:eacl17}, or active learning~\cite{woodward:arxiv2017,padmakumar:emnlp18}. 
The primary contributions of this work are introducing a dialog task that combines both clarification and active learning, and learning a corresponding dialog policy for this setting that outperforms a static baseline policy.
Specifically, we train a hierarchical dialog policy to jointly learn to choose clarification and active learning queries in interactive image retrieval for a fashion domain.

\begin{figure}[t]
    \centering
    \includegraphics[width=0.45\textwidth]{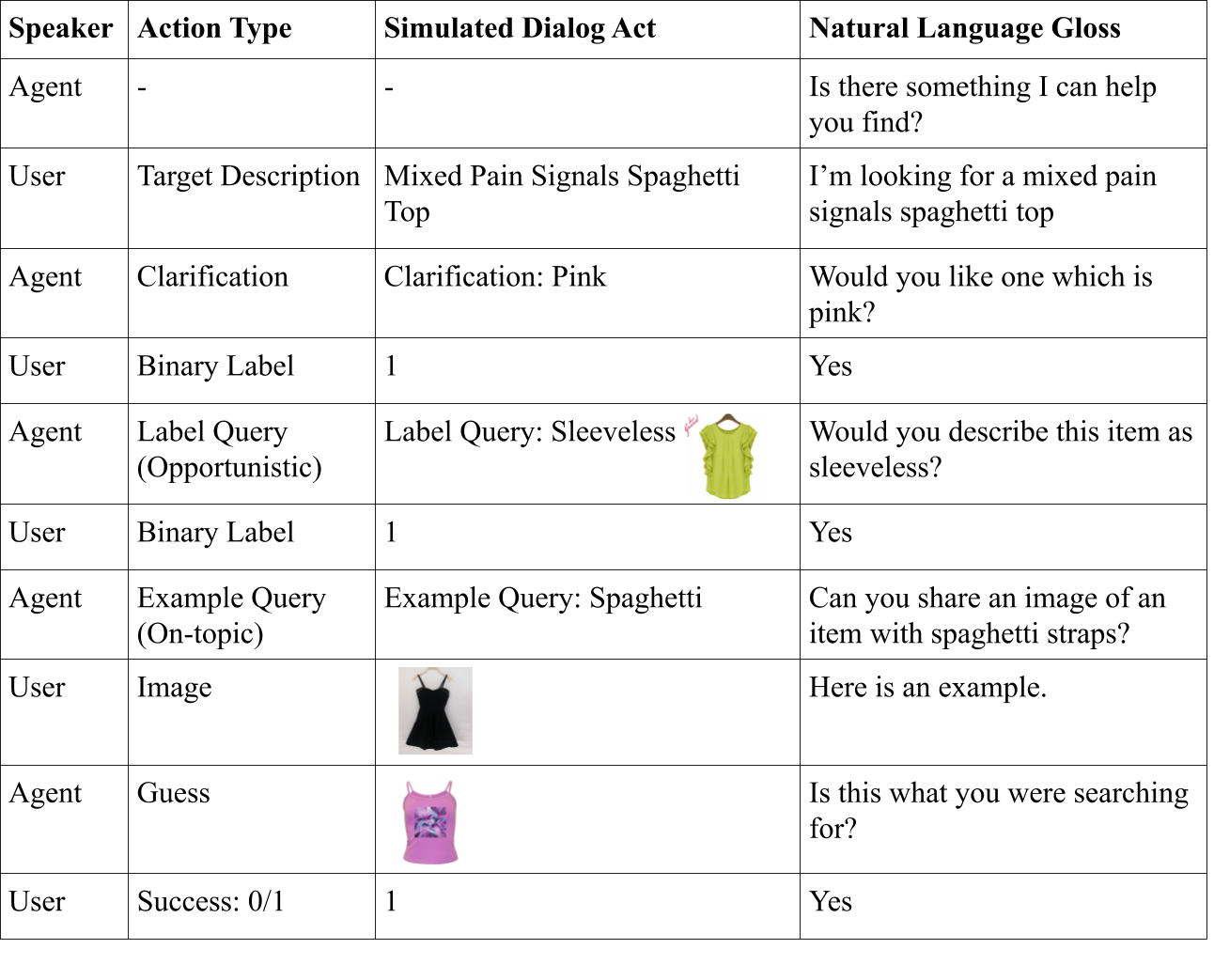}
    \caption{A stylized sample interaction. In this work, we use simulated dialog acts but the natural language glosses represent how such a dialog could look in an end application.}
    \label{fig:dialog}
\end{figure}

A sample interaction is shown in Figure \ref{fig:dialog}. We consider an application where a dialog system is combined with a retrieval system to help a customer find an article of clothing. Instead of just showing a large number of retrieved results, the dialog system attempts to use clarifications to refine the search query, and active learning questions to obtain labelled examples for novel concepts unseen during training.

Task-oriented dialog often requires the system to identify one or more user goals using a slot-filling model~\cite{young:procieee13}. 
These systems learn to choose between a set of clarification questions that confirm or acquire the value of various slots. 
However, for tasks such as natural language image retrieval, it is difficult to extend the slot-filling paradigm for clarification, as there is no standard set of slots into which descriptions of images can be divided.
Also, learned models are needed to identify aspects such as objects or attributes, which are difficult to pre-enumerate.

Some tasks such as GuessWhat?!~\cite{de:cvpr2017} or discriminative question generation~\cite{li:iccv17} allow the system to ask unconstrained natural language clarification questions. 
However they require specially designed models to ensure that learned questions actually decrease the search space~\cite{lee:vigil18,zhang:eccv18}. 
Such open ended questions are also difficult to answer in simulation, which is often necessary for learning good dialog policies.
Hence, in these tasks, the system often learns to ask ``easy'' questions that can be reliably answered by a learned answering module~\cite{zhu:arxiv17}.

In this work, we explore a middle-ground approach with a form of attribute-based clarification~\cite{farhadi:cvpr09}. 
We use the term ``attribute'' to refer to a mix of concepts including categories such as ``shirt'' or ``dress'', more conventional attributes such as colors, and domain specific attributes such as ``sleeveless'' and ``V-neck''.
Although we work with a dataset that contains a fixed set of attributes annotated for each image, we simulate the setting where novel visual attributes are encountered at test time. 

Dialog interaction can also be used to improve an underlying model using Opportunistic Active Learning  (OAL) \cite{padmakumar:emnlp18}. 
Active learning allows a system to identify unlabeled examples which, if labeled, are most likely to improve the underlying model. 
OAL \cite{thomason:corl17} incorporates such queries into an interactive task in which an agent may ask users questions that are irrelevant to the current dialog interaction to improve performance in future dialog interactions. 
Opportunistic queries are more expensive than traditional active learning queries as they may distract from the task at hand, but they can allow the system to perform more effective lifelong learning. 
Such queries have been shown to improve performance in interactive object retrieval~\cite{padmakumar:emnlp18}. 
However, this, and other works in reinforcement learning (RL) of policies for active learning~\cite{fang:emnlp2017} do not account for the presence of other interactive actions such as clarification.

We present a dialog task that combines natural language image retrieval with {\it both} OAL {\it and} attribute-based clarification.
We then learn a hierarchical dialog policy that jointly learns to choose both appropriate clarification and active learning questions in a setting containing both uncertain visual classifiers and novel concepts not seen during training. 
We observe that in our challenging setup, it is necessary to jointly learn dialog policies for choosing clarification and active learning questions to improve performance over employing one-shot retrieval with no interaction.

\section{Related Work}

Slot-filling style clarifications~\cite{young:procieee13} have been shown to be useful for a variety of domains including restaurant recommendation~\cite{williams:dd16}, restaurant reservation~\cite{bordes:iclr17}, movie recommendation and question answering~\cite{dodge:iclr16}, issuing commands to robots~\cite{deits:jhri13,thomason:ijcai15} and converting natural language instructions to code~\cite{chaurasia:ijcnlp17}.
Other tasks such as GuessWhat?!~\cite{de:cvpr2017}, playing 20 questions~\cite{hu:emnlp2018}, relative captioning~\cite{guo:nips18} and discriminative question generation~\cite{li:iccv17} enable very open-ended clarification.
Some works bring the setup closer to human-human conversations by allowing speakers to interrupt each other~\cite{manuvinakurike:acl17}.
In this work, we take an intermediate approach that allows finer-grained clarification than slot filling, but constrained so that reasonably accurate answers can be provided in simulation. 

Most of the above works learn dialog policies for clarification using RL~\cite{padmakumar:eacl17,Wen:naacl16,strub:ijcai17,hu:emnlp2018}. 
Some use information from clarifications to improve the underlying language-understanding model~\cite{thomason:ijcai15,padmakumar:eacl17}.
Such improvement is implicit in end-to-end dialog systems~\cite{Wen:naacl16,de:cvpr2017,hu:emnlp2018}. 
Instead, we use explicit active learning to improve the underlying perceptual model used for language grounding.
Some previous work uses visual attributes for clarification~\cite{dindo:rsj10,parde:ijcai15}, but they do not use this information to improve the underlying language understanding model.
There is also prior work on using active learning to elicit better user feedback to learn a better reward function to optimize dialog policies for task oriented dialog~\cite{su:acl16}. 
This direction is complementary to our work which is aimed at using active learning to improve the underlying language understanding model.

Hierarchical dialog policies have been designed for multi-task dialog systems where a top level policy alternates between subtasks such as booking a hotel or a flight, and a lower level policy that selects primitive dialog actions to complete the subtasks~\cite{peng:emnlp17,budzianowski:sigdial17}. 
Some hierarchical techniques such as feudal RL can enable dialog systems to scale to handle domains with a large number of slots~\cite{casanueva:naacl18}.
More related is \citep{zhang:sigdial18}, who design a hierarchical policy for visual dialog that has a low level policy for choosing clarification questions, and a high level decision policy to choose between clarification and guessing. Our work can be considered an extension of this framework that additionally accounts for active learning by including an additional low level policy for choosing active learning queries, and expanding the top level decision policy to choose between clarification, active learning as well as guessing.

Active learning has traditionally used  hand-coded sample-selection metrics, such as uncertainty sampling \cite{settles:2010}. 
Recent work on active learning in dialog/RL setups include using slot-filling style active learning questions to learn new names for known perceptual concepts~\cite{yu:robonlp17}, sequentially identify exampled to be labeled for a static task~\cite{fang:emnlp2017}, deciding between predicting a label for a specific example or requesting for it to be labelled~\cite{woodward:arxiv2017}, and jointly learning a data selection heuristic, data representation, and prediction function~\cite{bachman:icml17}. 
However, most of these (except ~\cite{woodward:arxiv2017}) do not involve a trade-off between active learning and task completion. 
None of them incorporate clarification questions. 

Most similar to our work is \cite{padmakumar:eacl17} which concerns learning a policy to trade-off opportunistic active learning questions to improve classifiers, and using these to ground natural-language descriptions of objects. 
However, instead of assuming a cold-start condition where the system cannot initially ground any descriptions before asking queries, we consider a warm-start condition closer to most real-world scenarios. 
We use a pretrained classifier and expect active learning to primarily aid generalization to novel concepts not seen during training.
We also extend the task to include clarification questions.     

Also related is work on interactive image retrieval such as allowing a user to mark relevant and irrelevant results~\cite{nastar:cvpr98,tieu:ijcv04}, which acts as a form of clarification. Recent works allow users to provide additional feedback using language to refine search results~\cite{guo:nips18,bhattacharya:mmr19,saha:aaai18}.
These directions are complementary to our work and could potentially be combined with it in the future. 

\section{Task Setup}

We consider an interactive task of retrieving an image of a product based on a natural-language description. Given a set of candidate images and a description, the goal is to identify the image being referred to. 
Before trying to identify the target, the system can ask a combination of both clarification and active learning questions. 
The goal is to maximize the number of correct product identifications across interactions, while also keeping dialogs as short as possible. 

Since we want to ensure that active learning questions are used to learn a generalizable classifier, we follow the setup of ~\cite{padmakumar:emnlp18} and in each interaction we present the system with two sets of images:
\begin{itemize}
    \item An active test set $\mathbf{I}^{Te}$ consisting of the candidate images to which the description could refer.
    \item An active training set $\mathbf{I}^{Tr}$ which is the set of images that can be queried for active learning.
\end{itemize}
It is also presented with a description of the target image. 
Before attempting to identify the target, the system can ask clarification or active learning questions.
We assume the system has access to a set of attributes $W$ that can be used in natural language descriptions of products. 
Given these attributes, the types of questions the system can ask are as follows (see Figure \ref{fig:dialog} for examples of each):
\begin{itemize}
    \item Clarification query - A yes/no query about whether an attribute $w \in W$ is applicable to the target. 
    \item Label query: A yes/no query about whether an attribute $w \in W$ is applicable to a specific image $i$ in the active training set $\mathbf{I}^{Tr}$. 
    \item Example query: Ask for a positive example in the active training set $\mathbf{I}^{Tr}$ for an attribute $w \in W$.
\end{itemize}

The dialog ends when the system makes a guess about the identity of the target, and is considered successful if it is correct. 
As in ~\cite{padmakumar:emnlp18}, we allow label and example queries that are either \textit{on-topic} (queries about attributes in the current description) or \textit{opportunistic}  (queries that are not relevant to the current description but may be useful for future interactions), which have been shown to help interactive object retrieval~\cite{thomason:corl17} (see Figure \ref{fig:dialog} for examples of each). 


\section{Methodology}

\subsection{Visual Attribute Classifier}
\label{ssec:classifier}

\begin{figure*}[t]
    \centering
    \includegraphics[width=0.8\textwidth]{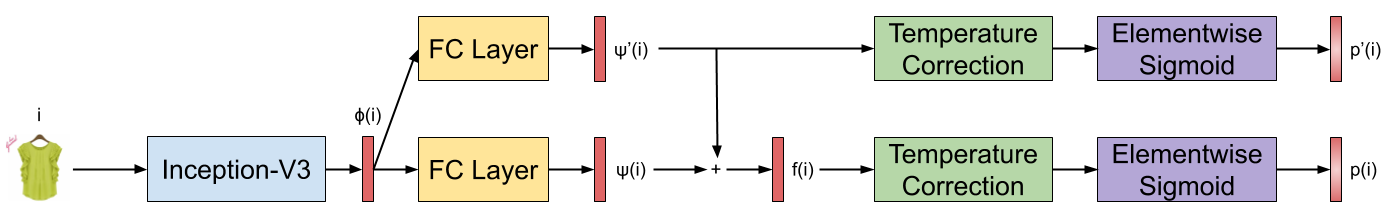}
    \caption{Visual Attribute Classifier}
    \label{fig:classifier}
    \vspace{-10pt}
\end{figure*}

We train a multilabel classifier for predicting visual attributes given an image. 
The network structure for the classifier is shown in Figure \ref{fig:classifier}.
We extract features $\phi(i)$ for the images using the penultimate layer of an Inception-V3 network~\cite{szegedy:cvpr16} pretrained on ImageNet~\cite{russakovsky:ijcv2015}.
These are passed through two separate fully connected (FC) layers with ReLU activations, that are summed to produce the final representation $f(i)$ used for classification. 
This is converted into per-class probabilities $p(i)$ using a sigmoid layer with temperature correction~\cite{guo:icml17}. 
We obtain another set of per-class probabilities $p'(i)$ by passing the one of the intermediate representations $\psi'(i)$ through a sigmoid layer with temperature correction.
Mathematically, given features $\phi(i)$ for image $i$, we have:
\begin{equation*}
\begin{aligned}[l]
\psi(i) & = ReLU(w^T\phi(i) + b) \\
\psi'(i) & = ReLU(w'^T\phi(i) + b') \\
f(i) & = \psi(i) + \psi'(i) \\
\end{aligned}
\hspace{25pt}
\begin{aligned}[l]
p(i) & = \sigma(f(i) \odot \frac{1}{\tau}) \\
p'(i) & = \sigma(\psi'(i) \odot \frac{1}{\tau'})
\end{aligned}
\end{equation*}
where $w$, $w'$, $\tau$ and $\tau'$ are learned vectors and $b$ and $b'$ are learned biases.

We train the network using a loss function that combines cross-entropy loss on $p(i)$ over all examples with the cross entropy loss over $p'(i)$ only for positive labels. That is,
\begin{eqnarray}
L & = & (1 - \lambda)\sum_i{y_i\log p(i) + (1 - y_i)\log(1 - p(i))} \nonumber\\
&& + \lambda\sum_i{y_i\log p'(i)} \nonumber
\end{eqnarray}
where $y_i$ is the label vector for image $i$.
This forces part of the network to focus on positive examples for each class (attribute). 
This is required because we use a heavily imbalanced dataset where most attributes have very few positive examples. 
We find this more effective than a standard weighted cross entropy loss, and the results in this paper use $\lambda = 0.9$.
We also maintain a validation set of images labeled with attributes, that can be extended using active learning queries. 
Using this, we can estimate per-attribute precision, recall and F1.
These metrics are used for tuning classifier hyperparameters and for dialog policy learning. More details about the classifier design are included in appendix \ref{sec:classifier_appendix}.

\subsection{Grounding Model}
\label{ssec:grounding}
We assume that a description is a conjunction of attributes, and use string-matching heuristics  to determine the set of attributes referenced by the natural language description. 
Let the subset of attributes in the description $d$ be $W_d \subseteq W$.

Suppose we additionally obtain from clarifications that attributes $W_p \subseteq W$ apply to the target image, and attributes $W_n \subseteq W$ do not apply, assuming independence of attributes, the probability that $i$ is the target image, $b(i)$ is:
\begin{equation}
    b(i) \propto \prod_{w \in W_d}{p_w(i)}\prod_{w \in W_p}{p_w(i)}\prod_{w \in W_n}{(1 - p_w(i))}
    \label{eqn:belief}
\end{equation}


At any stage, the best guess the system can make is the image with max belief, that is:
\begin{equation}
    i_{guess} = \mbox{argmax}_{i \in \mathbf{I}^{Te}}b(i)
    \label{eqn:guess}
\end{equation}

Also, we estimate the information gain $J$ of a clarification $q \in W$ as follows. This is based on the formulation used in ~\cite{lee:vigil18} but we additionally make a Markov assumption (details in appendix \ref{sec:info_gain}).
$$
J(q) = \sum_{i \in \mathbf{I}^{Te}}\sum_{a \in \{0, 1\}} b(i) P(a | q, i) \ln{\left( \frac{P(a | q, i)}{P(a|q)} \right)}
$$
where $P(1|q, i) = p_q(i)$ and $P(0|q, i) = 1 - p_q(i)$  and $P(a|q) = \sum_i{b(i)P(a | q, i)}$.

\subsection{MDP Formulation}
\label{sec:mdp}
We model each interaction as an episode in a Markov Decision Process (MDP) where the state consists of the images in the active training and test sets, the attributes mentioned in the target description, the current parameters of the classifier, and the set of queries asked and their responses. 
At each state, the agent has the following available actions:
\begin{itemize}
    \item A special action for guessing -- the image is chosen using Equation \ref{eqn:guess}. 
    \item One clarification query per attribute.
    \item A set of actions corresponding to possible active learning queries -- one example query per attribute and one label query for each pair $(w, i)$ for $w \in W$, $i \in \mathbf{I}^{Tr}$. 
\end{itemize}

We do not allow actions to be repeated.
We learn a hierarchical dialog policy composed of 3 parts -- clarification and active learning policies to respectively choose the best clarification and active learning query in the current state, and a decision policy to choose between clarification, active learning, and guessing.
An episode ends either when the guess action is chosen or a dialog length limit is reached, at which point the system is forced to make a guess. 
If the episode ends with a correct guess, the agent gets a large positive reward. Otherwise it gets a large negative reward at the end of the episode.
Additionally, we use a small negative reward for each query to encourage shorter dialogs. 
In our experiments, we treat these rewards as tunable hyperparameters.  

\subsection{Policy Learning}

We experimented with using both Q-learning and A3C~\cite{mnih:icml16} for policy learning, both  trained to maximize the discounted reward. Since the classifier has a large number of parameters, it is necessary to extract task-relevant features to represent state-action pairs. The features provided to each policy need to capture information from the current state that enable the system to identify useful clarifications and active learning queries, and trade off between these and guessing. The features used include:

\subsection{Clarification policy features}
\label{ssec:clarification_features}

\begin{itemize}
    \item Metrics about the current beliefs $\{b(i) : i \in \mathbf{I}^{Te}\}$ and what they would be for each possible answer, if the question were asked:
    \begin{itemize}
        \item Entropy: A higher entropy suggests that the agent is more uncertain. A decrease in entropy could indicate a good clarification. 
        \item Top two highest beliefs and their difference: A high value of the maximum belief, or a high difference between the top two beliefs could indicate that the agent is more confident about its guess. An increase in these could indicate a good clarification.
        \item Difference between the maximum and average beliefs: A large difference suggests that the agent is more confident about its guess. An increase in these could indicate a good clarification.
    \end{itemize}
    \item Information gain of the query as calculated in section \ref{ssec:grounding}.
    \item Current F1 of the attribute associated with the query: The system is likely to make better clarifications using attributes with high predictive accuracy.
\end{itemize}

\subsection{Active learning policy features}
\label{ssec:active_learning_features}

\begin{itemize}
    \item Current F1 of the attribute associated with the query, since the system is likely to benefit more from improving an attribute whose current predictive accuracy is not high.
    \item Fraction of previous dialogs in which the attribute has been used, since it is beneficial to focus on frequently used attributes that will likely benefit future dialogs.
    \item Fraction of previous dialogs using the attribute that have been successful, since this suggests that the attribute may be modelled well enough already.
    \item Whether the query is off-topic (i.e. opportunistic), since this would not benefit the current dialog.
\end{itemize}
Additionally, for label queries we use the following features:
\begin{itemize}
    \item For query $(w, i)$, $\vert p_w(i) - 0.5 \vert$ as a measure of (un)certainty.
    \item Average cosine distance of the image to others in the dataset; this is motivated by density weighting to avoid selecting outliers.
    \item Fraction of k-nearest neighbors of the image that are unlabelled for this attribute, since a higher value suggests that the query could benefit multiple images.
\end{itemize}

\subsection{Decision policy features}
\label{ssec:decision_features}

\begin{itemize}
    \item Features of the current belief as in Sec. \ref{ssec:clarification_features}. These can help determine whether a guess is likely to be successful.
    \item Information gain of the best clarification action -- to decide the utility of the clarification.
    \item Margin from the best active learning query if it is a label query -- to decide the utility of the label query.
    \item F1 of attributes in clarification and active learning queries. High F1 is desirable for clarification and low F1 for active learning.
    \item Mean F1 of attributes in the description. A high value suggests that the belief is more reliable.
    \item Number of dialog turns completed.
\end{itemize}

\subsection{Baseline static policy}
\label{ssec:baseline}

As a baseline, we use an intuitive manually-designed static policy that is also hierarchical and was tailored to perform well in preliminary experiments.
The static clarification policy chooses the attribute (among those with $F1>0$) with maximum information gain, with ties broken using F1.
The static active learning policy has a fixed probability of choosing label queries and example queries. Uncertainty sampling is used to select the label query $(w, i)$ with minimum $\vert p_w(i) - 0.5 \vert$. An example query is chosen uniformly at random from the candidates. 
The decision policy initially chooses clarification if the information gain is above a minimum threshold, and the highest belief is below a confidence threshold. After a maximum number of clarifications, it chooses active learning until another threshold on the dialog length before guessing.

\section{Experimental Setup}
\subsection{Dataset}
To address a potential shopping application, we simulate dialogs using the iMaterialist Fashion Attribute data \cite{guo:iccvworkshop19}, consisting of images from the shopping site Wish\footnote{\url{https://www.wish.com/}}  annotated for a set of 228 attributes. 
We scraped product descriptions for the images in the train and validation splits of the dataset for which attribute annotations are publicly available.
After removing products whose images or descriptions were unavailable, we had 648,288 images with associated descriptions and attribute annotations. 

We create a new data split following the protocol of \cite{padmakumar:emnlp18} to ensure that the learned dialog policy generalizes to attributes not seen during policy training. 
We divided the data into 4 splits, \textit{policy\_pretrain}, \textit{policy\_train}, \textit{policy\_val} and \textit{policy\_test}, such that each contains images that have attributes for which positive examples are not present in earlier splits to increase the potential benefit of active learning.
While we did not explicitly try to ensure that clarifications were beneficial, we validated that if we chose clarifications using an oracle that tries every clarification and selects the one that maximally increases the belief of the target image, it is possible to obtain a retrieval success rate of 80-85\% without performing any active learning. 
Each of these is then split into subsets \textit{classifier\_training} and \textit{classifier\_test} by a uniform 60-40 split.
More details are available in appendix \ref{sec:data_split}.

The \textit{policy\_pretrain} data is used to pretrain the multi-class attribute classifier. We use its \textit{classifier\_training} subset of images for training and its \textit{classifier\_test} subset to tune hyperparameters. 
The \textit{policy\_train} data is then used to learn the dialog policy. The \textit{policy\_val} data is used to tune hyperparameters  as well as choose between RL algorithms. Finally, results are reported for the \textit{policy\_test} data. 

We  simulate dialogs as refinements of an initial retrieval based on the product description (details in appendix \ref{sec:retrieval}).
For the description of each image in the current \textit{classifier\_test} subset, we rank all other images in this subset according to a simplified version of the score in equation \ref{eqn:belief}.
From the images that get ranked within the top 1000 for their corresponding description, we sample target images for each interaction. The active test set for the interaction consists of the top 1000 images as ranked for that description. 
We randomly sample 1000 images from the appropriate \textit{classifier\_training} subset to form the active training set.

As in \citep{padmakumar:emnlp18}, we start the simulation of a dialog by providing the description of the target image to the agent.
The annotated attributes are then used to automatically answer its queries and assess dialog success.

\subsection{Experiment Phases}
We run dialogs in batches of 100 and update the classifier and policies at the end of each batch. This is followed by repeating the retrieval step for all descriptions in the \textit{classifier\_test} subset before choosing target images for the next batch of dialogs.
The experiment has the following phases:
\begin{itemize}
    \item Classifier pretraining: We pretrain the classifier using annotated attribute labels for images in the \textit{classifier\_training} subset of the \textit{policy\_pretrain} set. This ensures that we have some reasonable clarifications at the start of dialog policy learning. 
    \item Policy initialization: We initialize the dialog policy using experience collected using the baseline static policies (section \ref{ssec:baseline}) for the decision and active learning policies, and an oracle~\footnote{The oracle tries each candidate clarification and returns the one that maximally increases the belief of the target image.} to choose clarifications. This is done to speed up policy learning. The dialogs for this phase are sampled from the set of \textit{policy\_train} images.
    \item Policy training: This phase consists of training the policy using on-policy experience, with dialogs again sampled from the set of \textit{policy\_train} images.
    \item Policy testing: We reset the classifier to the state at the end of pretraining. This is done to ensure that any performance improvement seen during testing are due to queries made during the testing phase. This is needed both for fair comparison with the baseline and to confirm that the system can generalize to novel attributes not seen during {\it any} stage of training. Dialogs are sampled for this from the \textit{policy\_val} set for hyperparameter tuning and from the \textit{policy\_test} set for reported results.  
\end{itemize}

\section{Results and Discussion}

We initialize the policy with 4 batches of dialogs, followed by 4 batches of dialogs for the training phase, and 5 batches of dialogs in the testing phase. 
We compare the fully learned policy with hierarchical policies that consist of keeping one or more of the components static. 
We also compare the choice of Q-Learning or A3C~\cite{mnih:icml16} as the policy learning algorithm for each learned policy. 
Table \ref{tab:main_results} shows the performance in the final test batch of the best fully learned policy, as well as a selected subset of the baselines (all conditions are included in appendix \ref{sec:extended_results}). We evaluate policies on the fraction of successful episodes in the final test batch, and the average dialog length. 

\begin{table}[t]
\caption{Results from the final batch of the test phase.}
\label{tab:main_results}
\begin{tabular}{m{0.2\linewidth}m{0.11\linewidth}m{0.16\linewidth}C{0.14\linewidth}R{0.1\linewidth}L{0.001\linewidth}}
     \toprule
     Decision Policy Type & 
     Clarifi-cation Policy Type & 
     Active Learning Policy Type & 
     Fraction of Successful Dialogs & 
     \multicolumn{2}{L{0.14\linewidth}}{Average Dialog Length}  \\
     \midrule
     Q-Learning & A3C & A3C & \textbf{0.33} & 9.40 & \\
     Q-Learning & A3C & Static & 0.15 & 14.16 & \\
     Q-Learning & Static & A3C & 0.09 & 1.00 & \\
     Static & A3C & A3C & 0.27 & 20.00 & \\
     Static & Static & Static & 0.17 & 20.00 & \\
     \bottomrule
\end{tabular}
\end{table}

Ideally, we would like the system to have a high dialog success rate while having as low a dialog length as possible. 
We observe that using a learned policy for all three functions results in a significantly more successful dialog system (according to an unpaired Welch t-test with $p<0.05$) than most conditions in which one or more of the policies are static. 
The exception is the case when the decision policy is static and the clarification and active learning policies are learned, in which case the difference is not statistically significant.
The fully learned policy also uses significantly shorter dialogs than all conditions with a static decision policy. 
Some other conditions result in shorter dialogs, but these are unable to exploit the clarification and active learning actions enough to result in a success rate comparable to the fully learned policy. 

\begin{figure}[t]
\begin{subfigure}{.5\textwidth}
  \centering
  \includegraphics[width=.8\linewidth]{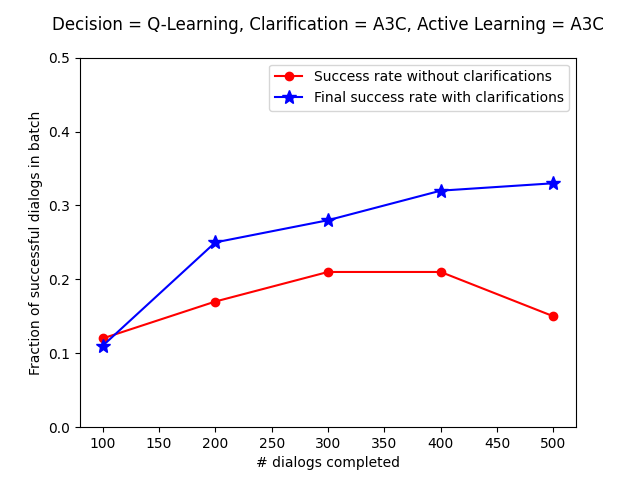}  
  \caption{Best Learned Policy}
  \label{sfig:d_Q__c_A2C__a_A2C}
\end{subfigure}
\begin{subfigure}{.5\textwidth}
  \centering
  \includegraphics[width=.8\linewidth]{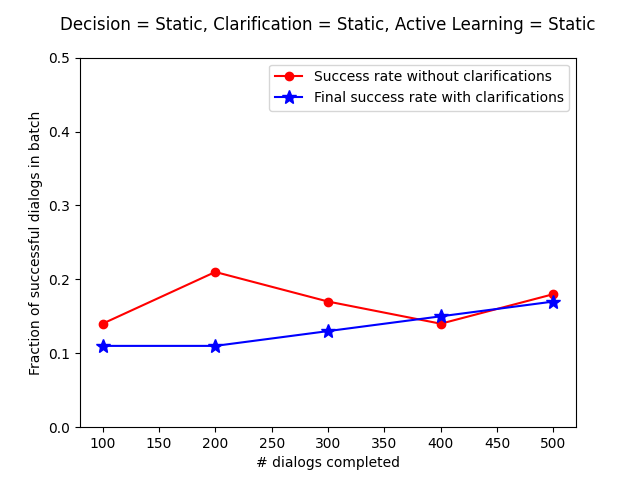}
  \caption{Baseline Static Policy}
  \label{sfig:d_static__c_static__a_static}
\end{subfigure}
\caption{Comparison of guess success rate with, and without clarifications across test batches.}
\label{fig:clarification_improvement}
\end{figure}

Figure \ref{fig:clarification_improvement} plots the success rate across test batches, and the expected success rate if the system was forced to guess without clarification, for the fully learned, and fully static policies. 
These can be used to determine the individual effects of clarification and active learning. 
A significant increase in the success rate without clarification from the first to the final test batch suggests that active learning by itself is improving the retrieval ability of the system. 
We do not find this to be the case, either for the fully static or the fully learned policy. 
This shows that neither the static active learning policy nor the learned active learning policy are able to improve performance in the absence of clarification.

A significant increase from the success rate without clarification to the success rate with clarification would demonstrate that clarification is beneficial for any particular test batch.
When performed on the first test batch, this demonstrates the effectiveness of clarifications alone, and when performed on the last test batch, this demonstrates the combined benefit of the clarification and active learning policies.

In the case of the fully static policy, we find that there is no statistically significant improvement, either in the expected initial success rate without clarifications, or in the final success rate, between the first and last test batch.
This suggests that neither the static active learning policy, nor its combination with the static clarification policy are capable of improving the system's performance.

However, in the case of the fully learned policy, we observe a statistically significant improvement in the final success rate, but not the initial success rate without clarifications. 
This suggests that while a learned active learning policy alone is not sufficient to improve the success rate, the {\it combination} of learned active learning and clarification policies {\it is} sufficient to improve the success rate.
We also observe that while the difference between the initial and final success rate is initially not significant, it increases across batches, and becomes significant in the last two batches.
This suggests that the clarification policy by itself is also insufficient for improvement, and the combination of the two is required to improve the system's success rate.

We believe that the reason for the relatively poor performance of the static clarification and active learning policies is that the classifier is not sufficiently accurate, and does not produce well calibrated probabilities, due to the heavy imbalance in the dataset. 
However, the learned policies are able to learn to properly adjust for this miscalibration.

\section{Human Evaluation}
\label{sec:human_eval}

We also compared our best learned policy and the baseline static policy using crowdworkers on Amazon Mechanical Turk. 
We had to make two important changes to our experiment setup for this evaluation.
We had to remove images from the iMaterialist dataset that may require content warnings when shown to crowdworkers using a manually curated list of 24 attributes.
We also found using pilot studies that workers on Amazon Mechanical Turk typically mentioned fewer of our annotated attributes than what is common in product descriptions, resulting in very low success rates from both policies.
To account for this, we made the retrieval task easier by having the dialog agent select the target image from a set of 100 random images, and sampled a subset of the attributes from the product description during training. 
We then retrained the learned policy for the modified task setup with 10 batches of initialization, 10 batches of training and 5 batches of testing in simulation.
We then used the final policy, and classifiers at the end of the test phase in interactions with users on Amazon Mechanical Turk to evaluate how well the learned system transfers.
An image of the interface seen by the workers and some additional details are included in appendix \ref{sec:amt_interface}.

To minimize confusion, we presented each step of the dialog on a new page, and provided the users with visual examples of attributes used in the questions.
We required workers to have completed at least 1000 HITs and have at least a 95\% approval rate on their previous HITs, as well as complete a qualification task to demonstrate that they understood the types of questions used in our experiment.
We had 50 workers interact with each system tested.

The results are shown in Table \ref{tab:amt_results}. We report the fraction of successful dialogs (in which the system guesses the correct target image), and average dialog length.
Firstly, we observe that in this condition, the improvement in the learned policy in simulation is considerably higher than the static policy. However, its average dialog length does not decrease. Qualitatively, we observe that in the test phase, the learned policy initially has a low rate of clarification, and high rate of active learning, which is reversed as the dialogs progress. We also observe that clarification is significantly more beneficial in this setting.
We also observe a considerable drop in success rate for both the static and learned policy, in human interactions.
However, the learned policy remains more successful than the static policy, trending towards significance (p $<=$ 0.1) according to an unpaired Welch t-test.

\begin{table}[t]
\caption{Results of the static and new learned policies at the end of the test phase in simulation and in interactions on Amazon Mechanical Turk. Bold indicates a statistically significant improvement over the baseline (p $<$ 0.05) and italic indicates trending significance (p $<=$ 0.1) according to an unpaired Welch t-test.}
\label{tab:amt_results}
\begin{tabular}{m{0.1\linewidth}C{0.19\linewidth}C{0.19\linewidth}C{0.15\linewidth}C{0.15\linewidth}}     
\toprule
     Policy &
     Simulation -- Fraction of Successful Dialogs & 
     Simulation -- Average Dialog Length &
     AMT -- Fraction of Successful Dialogs &
     AMT -- Average Dialog Length  \\
     \midrule
Static & 0.23 & 20.0 & 0.06 & 19.16 \\
Learned & \textbf{0.65} & 20.0 & \textit{0.16} & 18.86 \\
     \bottomrule
\end{tabular}
\end{table}

We speculate that the drop in performance may be because of differences between the labels in the original dataset, and those provided by workers, making our classifiers less effective.
We believe this is because some attributes are difficult for crowdworkers to label based on visual information alone, and we observed disagreement even on attributes we expected to be simpler, such as colors.

\section{Conclusion}
We demonstrate how a combination of RL learned policies for choosing attribute-based clarification and active learning queries can be used to improve an interactive system that needs to retrieve images based on a natural language description, while encountering novel attributes at test time not seen during training.
Our experiments show that in challenging datasets where it is difficult to obtain an accurate attribute classifier, learned policies for choosing clarification and active learning queries outperform strong static baselines.
We further show that in this challenging setup, a combination of learned clarification and active learning policies is necessary to obtain improvement over directly performing retrieval without interaction. 

\section{Future Work}
\label{sec:future_work}
In future, we would like to verify these by experimenting with different datasets, underlying tasks and grounding models. 
Another important future direction is expanding to more general types of clarification and active learning questions, for example free form natural language questions, and active learning examples provided as task demonstrations instead of binary labels. 
Other directions we would like to explore are techniques to improve the sample efficiency of the active learning methods involved, as well as few-shot adaptation of better pretrained grounding models, to increase the gains from performing active learning.  

\section*{Acknowledgements}

We would like to thank Peter Stone, Joydeep Biswas and the UT Austin BWI group for helpful discussions. We would also like to thank the anonymous reviewers for their feedback. This work was supported by a Google Faculty Award received by Raymond J. Mooney and NSF NRI grants IIS-1925082 and IIS-1637736.

\section*{Broader Impact}

Natural language interfaces such as language-based search, and intelligent personal assistants have the potential to make various forms of technology ranging from mobile phones and computers, as well as robots or other machines such as ATMs or self-checkout counters more accessible and less intimidating to users who are unfamiliar or uncomfortable with other interfaces on such devices such as command shells, button based interfaces or changing visual user interfaces.
Spoken language interfaces can also be used to make such devices more accessible for the visually impaired or users who have difficulty with fine motor control.

However, the use of these interfaces do involve concerns over privacy and data security. This is especially the case with devices based on spoken language interfaces as they need to analyze every conversation for potential codewords~\cite{lackes:19}. Thus, users need to trust that these extraneous conversations will not be stored, or analyzed for other information. 
This is particularly problematic in environments such as hospitals or lawyer's officers where confidentiality is expected.

Another concern is that transactions on these devices may be triggered by casual conversation or voices on television~\cite{liptak:17}, that were not intended to activate the dialog system. 
A related concern is that the ambiguity of language or mistakes made by the system may trigger unintended actions. 
In most applications, these can be handled by setting up appropriate confirmation or cancellation procedures for sensitive actions.
Increased use of clarification steps before execution of an action may provide an additional opportunity for users to cancel such actions before they take place.

Using active learning, or any form of continuous learning with user data can make machine learning systems more useful due to increased exposure to the data distribution with which such systems need to operate in practice.
However, most machine learning algorithms assume that the input data is complete and correct, both of which may be violated by systems that train on user-generated data. 
It is also possible for such data to be biased in a variety of ways -- ranging from potential absence of representation or misrepresentation of some groups of people who do not use the system as frequently, to filter-bubble like effects when many users provide a few frequent examples as training data to the system~\cite{baeza:16}.
Explicit active learning questions also allow users to deliberately provide misinformation to machine learning systems.
Practical systems using active learning need to incorporate methods for handling noisy data, and need to have tests in place for undesirable learned biases.

\bibliography{References}

\appendix

\section{Complete Results}
\label{sec:extended_results}

\begin{table}[!h]
\caption{Unabridged results from the final batch of the test phase. $^*$ indicates the conditions whose performance is comparable to the best condition (in bold).}
\label{tab:extended_results}
\begin{tabular}{m{0.12\linewidth}m{0.11\linewidth}m{0.16\linewidth}C{0.14\linewidth}R{0.1\linewidth}L{0.001\linewidth}}
     \toprule
     Decision Policy Type & 
     Clarifi-cation Policy Type & 
     Active Learning Policy Type & 
     Fraction of Successful Dialogs & 
     \multicolumn{2}{C{0.14\linewidth}}{Average Dialog Length}  \\
     \midrule
Q-Learning & Q-Learning & Q-Learning & 0.12 & 15.45 & \\
Q-Learning & Q-Learning & A3C & 0.18 & 16.96 & \\
Q-Learning & Q-Learning & Static & 0.14 & 11.83 & \\
Q-Learning & A3C & Q-Learning & 0.06 & 1.0 & \\
Q-Learning & A3C & A3C & \textbf{0.33} & 9.4 & \\
Q-Learning & A3C & Static & 0.15 & 14.16 & \\
Q-Learning & Static & Q-Learning & 0.17 & 13.96 & \\
Q-Learning & Static & A3C & 0.09 & 1.0 & \\
Q-Learning & Static & Static & 0.17 & 3.81 & \\
A3C & Q-Learning & Q-Learning & 0.09 & 20.0 & \\
A3C & Q-Learning & A3C & 0.19 & 20.0 & \\
A3C & Q-Learning & Static & 0.17 & 20.0 & \\
A3C & A3C & Q-Learning & 0.13 & 20.0 & \\
A3C & A3C & A3C & 0.09 & 20.0 & \\
A3C & A3C & Static & 0.15 & 20.0 & \\
A3C & Static & Q-Learning & 0.12 & 20.0 & \\
A3C & Static & A3C & 0.13 & 20.0 & \\
A3C & Static & Static & 0.16 & 20.0 & \\
Static & Q-Learning & Q-Learning & $^*$0.29 & 20.0 & \\
Static & Q-Learning & A3C & $^*$0.24 & 20.0 & \\
Static & Q-Learning & Static & 0.1 & 20.0 & \\
Static & A3C & Q-Learning & $^*$0.24 & 20.0 & \\
Static & A3C & A3C & $^*$0.27 & 20.0 & \\
Static & A3C & Static & 0.14 & 20.0 & \\
Static & Static & Q-Learning & 0.15 & 20.0 & \\
Static & Static & A3C & 0.16 & 20.0 & \\
Static & Static & Static & 0.17 & 20.0 & \\
     \bottomrule
\end{tabular}
\end{table}

Here, we include the complete set of results of which Table \ref{tab:main_results} is a part. For each of the three tasks -- choosing clarification questions, choosing active learning questions and deciding between these and guessing, we compare the use of a static policy with policies learned using Q-Learning and A3C. 

We evaluate policies on the fraction of successful episodes in the final test batch, and the average dialog length. An episode is considered successful if it ends with the system guessing the correct target item. 
Ideally, we would like the system to have a high dialog success rate while having as low a dialog length as possible. 

We observe that the using a learned policy for all three functions, with A3C for choosing clarification and active learning queries, and Q-learning for deciding between clarification, active learning and guessing, results in a significantly more successful dialog system (according to an unpaired Welch t-test with p < 0.05) than most other conditions.
The exceptions are when the decision policy is static and the clarification and active learning policies are learned (marked by $^*$ in Table \ref{tab:extended_results}), in which the difference is not statistically significant.
However, the best fully learned policy uses significantly shorter dialogs than the other policies with a comparable success rate, thus making it overall more desirable. 

\section{Data Split}
\label{sec:data_split}

We divide the set of attributes into 4 subsets -- \textit{policy\_pretrain}, \textit{policy\_train}, \textit{policy\_val} and \textit{policy\_test}. 
Using these, we divide the images into 4 subsets as follows:
\begin{itemize}
    \item All images having a positive label for any of the attributes in \textit{policy\_test} subset form the \textit{policy\_test} set of images. 
    \item Of the remaining images, the images with a positive label for any attribute in the \textit{policy\_val} form the \textit{policy\_val} set of images. 
    \item Of the remaining images, the images with a positive label for any attribute in the \textit{policy\_train} form the \textit{policy\_train} set of images.
    \item The remaining images form the set of \textit{policy\_pretrain} images. 
\end{itemize}
This iterative procedure ensures that images in each of the \textit{policy\_train}, \textit{policy\_val} and \textit{policy\_test} result in the introduction of new attributes for which the classifier is not already trained.
The dataset has 228 attributes, of which 172 are present during initial pretraining, 13 are introduced in the dialog training phase, 21 are introduced during validation, and 22 are introduced in the test phase.

This increase the potential benefit of active learning. While we did not explicitly try to ensure that clarifications were beneficial, we validated that if we chose clarifications using an oracle that tries every clarification and selects the one that maximally increases the belief of the target image, it is possible to obtain a retrieval success rate of 80-85\% without performing any active learning.
The data split will be included in the code release.

\section{Initial Retrieval}
\label{sec:retrieval}

We wish to simulate dialogs as refinements of an initial retrieval based on the product description.
At the start of each batch of interactions, for each description corresponding to an image in the current \textit{classifier\_test} subset, we rank all images in this subset according to a variant of the score in Equation \ref{eqn:belief}. 
Instead of directly using classifier probabilities, we threshold the probabilities $p_w(i)$ to obtain decisions $d_w(i)$. 
The threshold for each attribute is chosen to maximize the F1 score for that attribute on the current set of validation images and labels.
This is initially the \textit{classifier\_test} subset of the \textit{policy\_pretrain} set, and gets expanded with a fraction of the labels obtained using active learning queries. 
This F1 score is also used in the baseline static policy (section \ref{ssec:baseline}) and features for the learned dialog policies (sections \ref{ssec:clarification_features}, \ref{ssec:active_learning_features} and \ref{ssec:decision_features}).

Also, while the initial belief in the dialogs (Equation \ref{eqn:belief}) only assumes that attributes mentioned in the description are positive for the target image, in the retrieval phase, we additionally assume that attributes not mentioned in the description are negative.
Then,
\begin{eqnarray}
    b(i) & = & \prod_{w \in W_d}{p_w(i)}\prod_{w \notin W_d}{(1 - p_w(i))} \\
    & \approx & \prod_{w : d_w(i) = 1}{c_1}\prod_{w : d_w(i) = 0}{c_2} \nonumber\\
    s(i) & = & \log{b(i)} \\
    & = & c_1 \vert \{w : d_w(i) = 1\} \vert + c_2  \vert \{w : d_w(i) = 0\} \vert \nonumber
\end{eqnarray}
We use the score $s(i)$ to rank images, where $c_1$ and $c_2$ are hyperparameters tuned on the \textit{classifier\_test} subset of the \textit{policy\_pretrain} set. Our reported results use $c_1 = 0.9$ and $c_2 = 0.1$.

\section{Policy Representation and Learning}
\label{sec:policy_rep}
At any state $s$, the agent can take one of the following actions:
\begin{itemize}
    \item A special action $a_g$ for guessing -- the image is chosen using Equation \ref{eqn:guess}. 
    \item A set of clarification actions $A_C$ -- one for each attribute.
    \item A set of actions $A_L$ corresponding to possible active learning queries -- one example query per attribute and one label query corresponding to each pair $(w, i)$ for $w \in W$, $i \in \mathbf{I}^{Tr}$. 
\end{itemize}
We learn a hierarchical dialog policy that consists of three parts as described below. For each policy, we obtain a feature representation of a state-action pair, outlined in sections \ref{ssec:clarification_features}, \ref{ssec:active_learning_features} and \ref{ssec:decision_features}, and discussed in further detail below. 
\begin{itemize}
    \item A clarification policy $\pi_C$ to choose the best possible clarification action $a_C^* \in A_C$ in the current state, using features $\phi_C(s, a_C)$ for action $a_C$ in state $s$.
    \item An active learning policy $\pi_{AL}$ to choose the best possible active learning query $a_{AL}^* \in A_{AL}$ in the current state, using features $\phi_{AL}(s, a_{AL})$ for action $a_{AL}$ in state $s$. We reduce the action space of the active learning policy to one example query action per attribute, and one label query action per attribute corresponding to the image with probability closest to 0.5 for that attribute ($argmin_{i \in \mathbf{I}^{Tr}}{\vert p_w(i) - 0.5 \vert}$).
    \item A decision policy $\pi_D$ that chooses between $a_g$, $a_C^*$ and $a_{AL}^*$, using features $\phi_D(s, a)$ for action $a$ in state $s$.
\end{itemize}

We experiment with using both Q-learning and A3C~\cite{mnih:icml16} for policy learning. We use the same model structure for all three policies but no shared parameters. In the following discussion about the model structure for the policy, we will refer to state-action features $\phi(s, a)$ and policy $\pi(s, a)$, which is intended to represent the appropriate input and output for each policy. 

For Q-learning we use a single-layer neural network with hidden layer size 100, whose input is the feature vector $\pi(s, a)$, and output is the Q-value $Q_\pi(s, a)$ of action $a$ in state $s$ under policy $\pi$. 
Suppose action $a$ is taken in state $s$ resulting in reward $r$ and next state $s'$, we update the network with new targets:
$$
Q_\pi(s, a) = r + \gamma * \max_{a'}{Q_\pi(s', a')}
$$
In the policy training phase, we choose $\epsilon$-greedy actions with $\epsilon = 0.1$ and in the policy validation and testing phases, we choose actions greedily. We use $\gamma = 1.0$.

For A3C, as a critic, we use a network similar to Q-learning, predicting $Q_\pi(s, a)$. The actor uses a policy representation:
$$
\pi_\theta(s, a) = \frac{e^{\theta^T \phi(s, a)}}{\sum_{a'}{e^{\theta^T \phi(s, a')}}}
$$
where $\theta$ is a learned parameter vector. 
Suppose action $a$ is taken in state $s$ resulting in reward $r$ and next state $s'$, the critic network is updated similar to Q-learning and the actor weights are updated as:
\begin{eqnarray}
&& \theta \leftarrow \theta + \alpha \nabla_\theta{\log{\pi_\theta(s, a)A_\pi(s, a)}}  \nonumber\\    
\mbox{where} && A_\pi(s, a) = r + \gamma * V_\pi(s') - V_\pi(s) \nonumber\\    
\mbox{where} && V_\pi(s) = \sum_a{\pi_\theta(s, a)Q_\pi(s, a)} \nonumber
\end{eqnarray}
where $Q_\pi(s, a)$ is the estimate from the critic network.
We use $\gamma = 1.0$ and $\alpha = 0.01$.

We now include a fuller description of the features used by each policy.
For the clarification policy $\pi_C$, we need to extract features of a state-action pair, that is, features describing a candidate clarification given the current state. The features should provide information that could be used to identify useful clarifications. We use the following features -
\begin{itemize}
    \item Metrics about the current beliefs $\{b(i) : i \in O^{te}_A\}$, as calculated in section \ref{ssec:grounding}, and what they would be updated to for each possible answer, if the clarification question were asked:
    \begin{itemize}
        \item Entropy = $-\sum_{i \in \mathbf{I}^{Te}}{b(i)\log{b(i)}}$: A higher entropy suggests that the agent is more uncertain. A decrease in entropy could indicate a good clarification. 
        \item Maximum belief, $b^1 = \max_{i \in \mathbf{I}^{Te}}{b(i)}$ - A high value of the maximum belief could indicate that the agent is more confident about its guess. Let the image corresponding to this belief be $\hat{i}$. An increase in this could indicate a good clarification.
        \item Second highest belief $b^2 = \max_{i \in \mathbf{I}^{Te} - \{\hat{i}\}}{b(i)}$ - A low value of the second highest belief suggests that the belief is concentrated in the maximum value and could indicate that the agent is more confident about its guess. A decrease in this could indicate a good clarification.
        \item Difference between the top two beliefs = $b^1 - b^2$: A high difference between the top two beliefs could indicate that the agent is more confident about its guess. An increase in this could indicate a good clarification.
        \item Average belief, $\Bar{b} = \frac{1}{\vert \mathbf{I}^{Te} \vert}\sum_{i \in \mathbf{I}^{Te}}{b(i)}$ - A large difference between the maximum and average beliefs could indicate that the agent is more confident about its guess. A decrease in this could indicate a good clarification.
        \item Difference between the maximum and average beliefs = $b^1 - \Bar{b}$: A large difference suggests that the agent is more confident about its guess. An increase in this could indicate a good clarification.
    \end{itemize}
    \item Information gain of the query as calculated in section \ref{sec:info_gain}.
    \item Current F1 of the attribute associated with the query: The system is likely to make better clarifications using attributes with high predictive accuracy.
\end{itemize}

For the active learning policy $\pi_{AL}$, we need to extract features describing a candidate label or example query given the current state. We use the following features:
\begin{itemize}
    \item Current F1 of the attribute associated with the query, since the system is likely to benefit more from improving an attribute whose current predictive accuracy is not high.
    \item Fraction of previous dialogs in which the attribute has been used, since it is beneficial to focus on frequently used attributes that will likely benefit future dialogs.
    \item Fraction of previous dialogs using the attribute that have been successful, since this suggests that the attribute may be modelled well enough already.
    \item Whether the query is off-topic (i.e. opportunistic), since this would not benefit the current dialog.
\end{itemize}
Additionally in label queries,
\begin{itemize}
    \item For query $(w, i)$, $\vert p_w(i) - 0.5 \vert$ as a measure of (un)certainty.
    \item Average cosine distance of the image to others in the dataset; this is motivated by density weighting to avoid selecting outliers.
    \item Fraction of k-nearest neighbors of the image that are unlabelled for this attribute, since a higher value suggests that the query could benefit multiple images.
\end{itemize}

For the decision policy $\pi_D$, we need features of the current state that allow the system to trade-off between clarification, active learning, and guessing. 
\begin{itemize}
    \item Features of the current belief used in clarification features -- These can be used to determine whether a guess is likely to be successful.
    \item Information gain of the best clarification action -- to decide the utility of the clarification.
    \item Margin from the best active learning query if it is a label query -- to decide the utility of the label query.
    \item F1 of attributes in clarification and active learning queries. High F1 is desirable for clarification and low F1 for active learning.
    \item Mean F1 of attributes in the description. A high value suggests that the belief is more reliable.
    \item Number of dialog turns completed.
\end{itemize}

\section{Estimation of Information Gain}
\label{sec:info_gain}

Our static baseline for choosing clarification questions is based on prior work in goal-oriented dialog that attempts to estimate the information gain of a clarification question~\cite{lee:vigil18}. In this setting, the agent asking questions needs to identify a target object among a set of candidate objects, and can ask clarification questions to help identify the target. 
Let $C$, $Q_t$ and $A_t$ be random variables corresponding to the target object, question in turn $t$ and answer in turn $t$ respectively, and $c$, $q_t$ and $a_t$ represent specific values of these variables. 
Then the information gain from asking question $q_t$, given previous questions $q_{1:t-1}$ and their answers $a_{1:t-1}$ is 
\begin{eqnarray}
&& I[C, A_t; q_t, a_{1:t-1}, q_{1:t-1}] \nonumber\\
& = & \sum_{a_t} \sum_c \left\{ p(c \vert a_{1:t-1}, q_{1:t-1})p(a_t \vert c, q_t, a_{1:t-1}, q_{1:t-1}) \right. \nonumber\\
&& \left. \times \ln{\frac{p(a_t \vert c, q_t, a_{1:t-1}, q_{1:t-1})}{p(a_t \vert q_t, a_{1:t-1}, q_{1:t-1})}} \right\} \nonumber
\end{eqnarray}
where

\begin{eqnarray}
&& p(a_t \vert q_t, a_{1:t-1}, q_{1:t-1}) \nonumber\\
& = & \sum_c{p(c \vert a_{1:t-1}, q_{1:t-1})p(a_t \vert c, q_t, a_{1:t-1}, q_{1:t-1})} \nonumber
\end{eqnarray}

In our case, possible targets $c$ correspond to possible images $i$. As in prior work~\cite{lee:vigil18}, $p(c \vert a_{1:t-1}, q_{1:t-1})$ corresponds to the estimated likelihood of target $c$ given the conversation history, which in our case is $b(i)$.
We also make an additional assumption that the answer to question $q_t$ depends only on the target image and not on prior questions and answers. Hence: 
$$
p(a_t \vert c, q_t, a_{1:t-1}, q_{1:t-1}) = p(a_t \vert c, q_t)
$$
which in our case is $P(a_t \vert q_t, i)$. Since our questions $q_t$ are attributes and the attribute classifier is expected to provide the probability that attribute $q$ is true for image $i$, we get $P(1|q, i) = p_q(i)$ and $P(0|q, i) = 1 - p_q(i)$. 
In practice, we observe that the classifier does not produce well-calibrated probabilities despite the use of temperature correction, and we believe that this contributes to the poor performance of the static clarification policy. 

Substituting these, we get information gain for question $q$, which we represent using $J(q)$, as:
$$
J(q) = \sum_{i \in \mathbf{I}^{Te}}\sum_{a \in \{0, 1\}} b(i) P(a | q, i) \ln{\left( \frac{P(a | q, i)}{\sum_i{b(i)P(a | q, i)}} \right)}
$$

\section{Classifier Design and Hyperparameters}
\label{sec:classifier_appendix}

For the attribute classifier, we initially experimented with alternate classifier designs such as binary SVMs using features extracted from Inception-V3 and fine-tuning Inception-V3 after altering the number of classes. We also experimented with alternate loss functions for fine-tuning Inception-V3, as well as the design in section \ref{ssec:classifier} such as weighted cross entropy, and a ranking loss that maximizes the difference between the predicted probabilities of positive and negative attributes. Additionally, we compared fine-tuning all layers of Inception-V3 with training/fine-tuning only the extra/final layers. 
We used Inception-V3 as the backbone network due to the results reported in the original paper~\cite{guo:iccvworkshop19}.

In our design, the branch producing p’ is expected to perform a function similar to reweighting of examples, in order to allow the classifier to handle a very imbalanced dataset. Since most attributes are negative for most examples, standard  neural network architectures overfit to predicting a 0 for every attribute for any input image. To avoid this, we add the p’ branch which only gets a loss for false negatives. Again, if this was independent, it could overfit by predicting everything to be positive, so we sum the two branches to obtain an effect that is mathematically similar to reweighting positive examples, but empirically worked better than reweighting for this dataset.

In contrast to the original paper~\cite{guo:iccvworkshop19}, we found that our particular network design and loss function were required for obtaining reasonable classifier performance. 
Additionally, we found that it was required to initialize Inception-V3 with weights pretrained on ImageNet and train only the new layers on the iMaterialist dataset.
These differences could be due to the differences in the data split. 
Our choice of data split results in many attributes always having a negative label during the training phase. 

We also found that it was sometimes possible to obtain increases in the multilabel F1 metric proposed in the original paper~\cite{guo:iccvworkshop19} without any improvement on per attribute F1. 
For example, it is possible to obtain a multilabel F1 of 36.0 on the original validation set by identifying the 13 attributes with the largest number of positive examples, always predicting 1 for these, and always predicting 0 for the other attributes. 
Hence, we used the average per-attribute F1 to choose the design and tune hyperparameters of the classifier.

To initialize the classifier, we train for 100 epochs with a batch size of 8,192 and using RMSProp for optimization. We start with a learning rate of 0.1 which is decayed exponentially with a decay rate of 0.9 every 400 steps. 
At the end of pretraining, the F1 scores of the classifier for attributes that get learned range from 0.39 - 0.94. 
Some attributes are not learned at this point, either because they are intended to be novel, and hence no positive examples have been seen, or because they are too heavily imbalanced.
To validate that these classifiers were reasonable to use for dialogs, we experimented with conducting dialogs using the baseline static dialog policy but choosing clarification questions using an oracle that examined all possible clarification questions and selected the one that maximally increased the belief of the correct target object.
This policy had a dialog success rate of 80-85\% (depending on hyperparameter settings) before any active learning. 
This is considered to be an oracle policy because it uses knowledge of the correct target object which would not be normally available to the dialog policy.

These will need to be learned from examples obtained during dialog interaction.
For updating the classifier in between dialog batches, we use a batch size of 128 and perform a single epoch over images for which the label of at least one attribute has been updated.  

\section{Generalization to Other Tasks}
\label{sec:generalization}

We believe our overall goal of combining clarification and active learning questions in dialog systems is an interesting direction in which we hope to see more work in the future. We consider our work to be an initial template for creating similar tasks in other dialog domains. 
Our framework assumes that active learning is relevant to the functionality of the dialog system. Hence, we assume that the dialog system is using a machine-learned model for some sort of language grounding. This is applicable for most visually grounded dialog, but can also be used in other situations. For example, a dialog system that is operating over a recommendation model can perform active learning by querying a user for ratings of past experiences, or general preferences. 
We also believe that our specific choice of attribute-based clarifications may be relevant for many situations where a slot can have a large number of values. For example, in a movie recommendation domain, we may have some well-catalogued attributes such as the director or the genre, but it may also be relevant to some other properties that are less likely to be annotated, such as “feel-good”.
We acknowledge that the exact style of clarification and active learning questions may need to be varied based on the task, and the underlying machine-learning model used for grounding. Our goal was to make an initial attempt at combining the functions of clarification and active learning in a single dialog task, and we considered binary questions for both functions to be a reasonable starting point. 
We would also like to acknowledge that the numerical gains from performing clarification and active learning may be less visible in some domains where the initial grounding model is very accurate and generalizable. In a practical system, we expect our work on clarification and active learning to handle edge cases and rarer cases in the tail of a Zipfian distribution that cannot all be covered by the initial training. For some types of grounding models, we acknowledge that this would require better active learning methods than those currently available. 

\section{Amazon Mechanical Turk Interface}
\label{sec:amt_interface}

\begin{figure}[!t]
    \centering
    \begin{subfigure}{\linewidth}
        \includegraphics[width=0.85\linewidth]{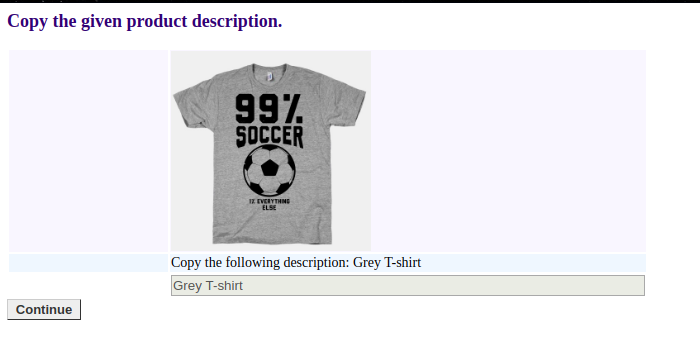}
        \subcaption{Step 1: Copy an example description.}
        \label{fig:amt_stepwise_example_description}
    \end{subfigure}
\medskip
    \begin{subfigure}{\linewidth}
        \includegraphics[width=0.85\linewidth]{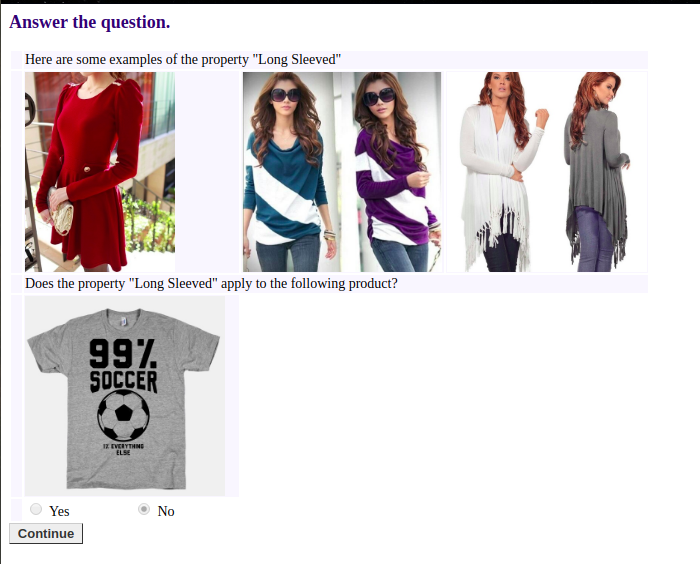}
        \subcaption{Step 2: Clarification Question.}
        \label{fig:amt_stepwise_clarification}
    \end{subfigure}
    \centering
    \begin{subfigure}{\linewidth}
        \includegraphics[width=0.9\textwidth]{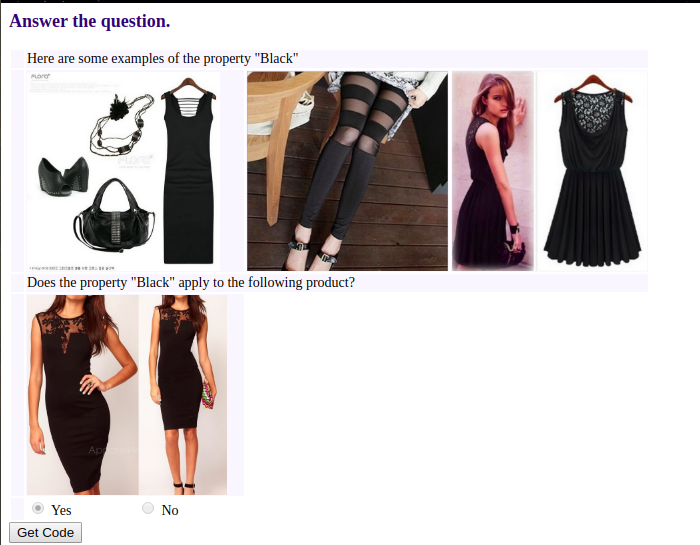}
        \subcaption{Step 3: Active Learning Question.}
        \label{fig:amt_stepwise_active_learning}
    \end{subfigure}
    \caption[]{Final qualification interface for Amazon Mechanical Turk.}
    \label{fig:amt_stepwise}
\end{figure}

We originally wanted to create an interface for workers that looked like a chat, but also enabled them to easily understand the task. 
However, a pilot study using such an interface showed that very few workers were successfully able to complete a qualification task that involved copying a given product description, and answering one clarification and one active learning question based on simple attributes. 
Hence, we changed the interface to provide one question per page, and allow the worker to move to the next page only after they had provided an answer to the question on the current page.
The resultant qualification task had 3 pages as shown in Figure \ref{fig:amt_stepwise}.
This change significantly increased the number of workers who qualified for our main experiment through this task.
Further work is needed to develop an appropriate user interface for this task that allows workers to interact in a more dialog-like fashion, while also not getting confused by the amount of information provided. 

Our goal was to choose relatively unambiguous clarification and active learning queries for the qualification task, and only allow users who correctly copied the description and provided the answers we expected to both queries to participate in our main experiment. 
We provided manually chosen example images for each attribute because while some attributes such as colors may be known to all workers, other such as ``Argyle'' or ``Herringbone'' may not be familiar to the average worker, but we believe can be identified given examples. 
We required workers to have completed at least 1000 HITs and have at least a 95\% approval rate on their previous HITs.

\begin{figure}[ht]
    \centering
    \includegraphics[width=0.9\linewidth]{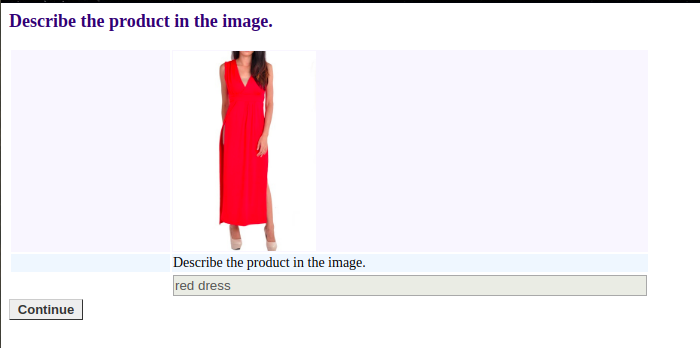}    
    \caption{Final dialog task on Amazon Mechanical Turk -- description page.}
    \label{fig:amt_stepwise_description}
\end{figure}

For the final experiment, we added a page at the start that showed the target image and asked users to describe the target image (Figure \ref{fig:amt_stepwise_description}). 
This was followed by pages similar to figures \ref{fig:amt_stepwise_clarification} and \ref{fig:amt_stepwise_active_learning} as relevant, based on the dialog system's responses.
Finally when the dialog system made a guess, we determined whether the system was successful based on whether the retrieved image was the true target image.

\section{Qualitative Examples and Discussion}
\label{sec:qualitative}

A sample interaction on AMT using the learned dialog policy is included in Table \ref{tab:amt_example}. 
While the agent is eventually successful, we can see that the interaction is fairly long and can be somewhat tedious.
In the more challenging simulated setup using in Table \ref{tab:extended_results}, many of the learned policies could on average to produce shorter dialogs. 
Further work needs to be done to easily identify hyperparameter settings that would translate these benefits when the task setup is changed. 

Also, in a real world application, we would expect active learning queries to be spread across a larger number of users so that individual users do not have to answer too many queries. 
Additionally, we would need improved few shot learning methods that can produce more improvement in a large pretrained classifier from a few active learning examples. 

\begin{figure}[ht]    
\centering
\includegraphics[width=0.8\linewidth]{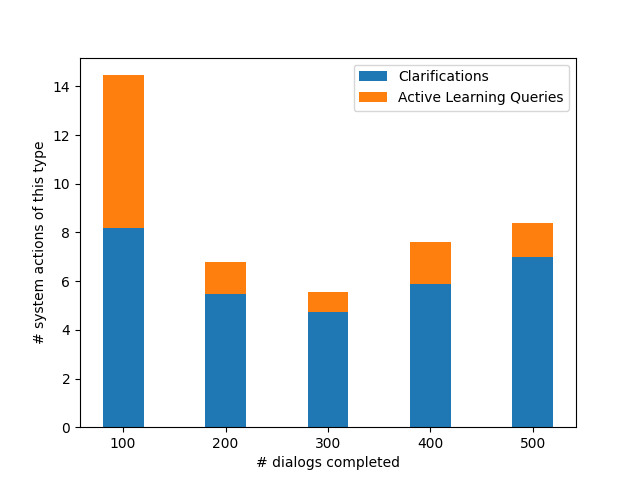}
\caption{Dialog turns of each type taken by the best learned policy in the test phase.}
\label{fig:question_split}
\end{figure}

We also examine the statistics of different dialog acts taken by the best learned policy to qualitatively assess how it differs from the static policy. The results are shown in Figure \ref{fig:question_split}. We observe that the average number of clarification questions remains relatively stable across test batches, whereas active learning questions are concentrated into the first test batch (when it is trying to learn about novel attributes), with far fewer active queries in later test batches. \\

\tablecaption{A sample interaction from Amazon Mechanical Turk.}
\label{tab:amt_example}
\tablefirsthead{\toprule}
\tablehead{%
\multicolumn{2}{c}%
{{\textit{Continued from previous column}}} \\
\toprule
}
\tabletail{%
\midrule \multicolumn{2}{r}{{\textit{Continued on next column}}} \\ \midrule}
\tablelasttail{%
\\\midrule
\multicolumn{2}{r}{{Concluded}} \\ \bottomrule}
\begin{xtabular}{L{0.1\linewidth}C{0.1\linewidth}C{0.1\linewidth}C{0.1\linewidth}C{0.7\linewidth}}
Agent & \multicolumn{4}{L{0.8\linewidth}}{What can I help you find?} \\
User & \multicolumn{4}{L{0.8\linewidth}}{Black Off-the-Shoulder Dress} \\
Agent & \multicolumn{4}{L{0.8\linewidth}}{Here are some examples of the property "Female"} \\
&
\begin{minipage}{.3\linewidth}
    \includegraphics[height=10mm]{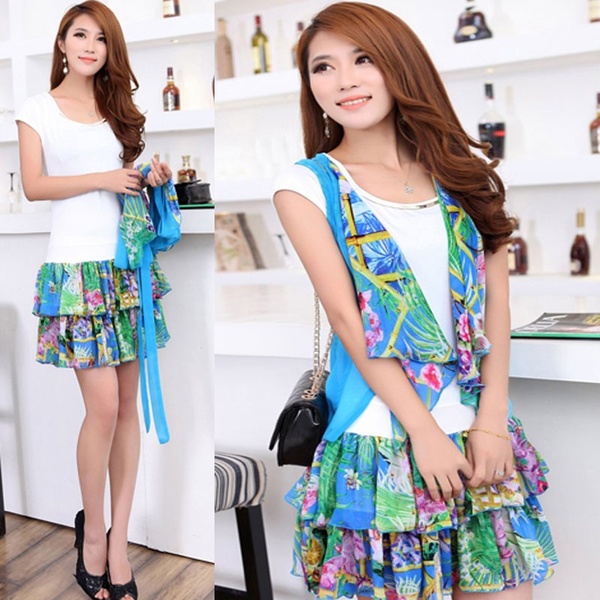}
\end{minipage}
&
\begin{minipage}{.3\linewidth}
    \includegraphics[height=10mm]{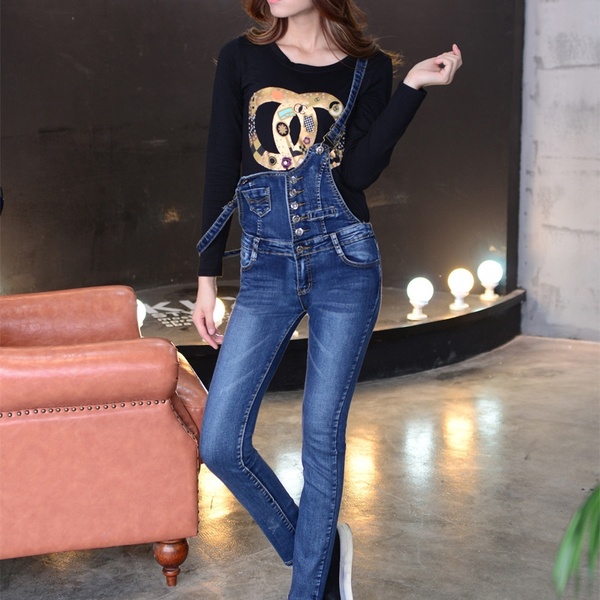}
\end{minipage}
&
\begin{minipage}{.3\linewidth}
    \includegraphics[height=10mm]{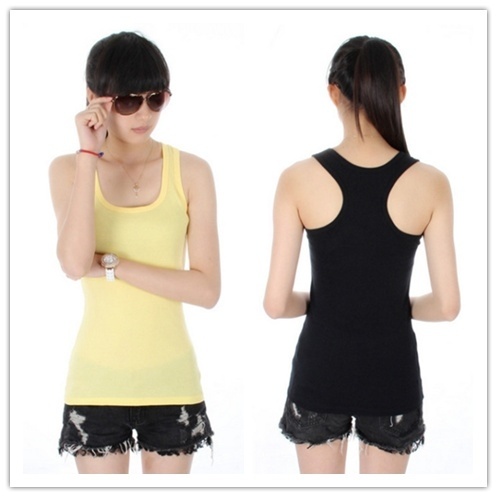}
\end{minipage} \\
Agent & \multicolumn{4}{L{0.8\linewidth}}{Does the property "Female" apply to the product you want?} \\
User & \multicolumn{4}{L{0.8\linewidth}}{Yes} \\
Agent & \multicolumn{4}{L{0.8\linewidth}}{Here are some examples of the property "Round Neck"} \\
&
\begin{minipage}{.3\linewidth}
    \includegraphics[height=10mm]{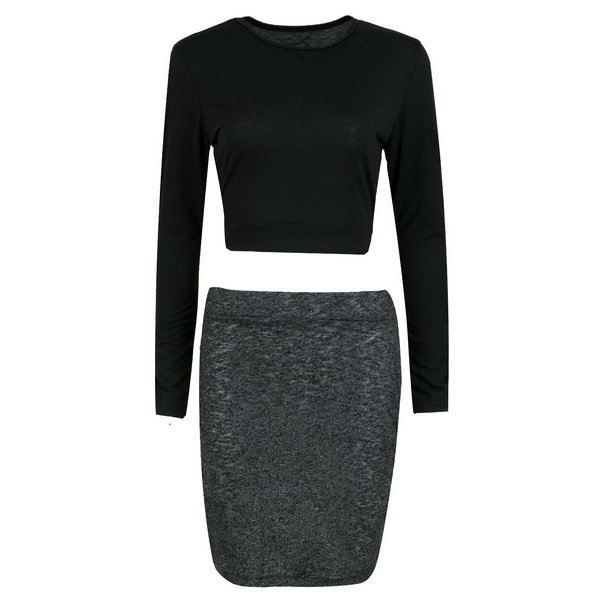}
\end{minipage}
&
\begin{minipage}{.3\linewidth}
    \includegraphics[height=10mm]{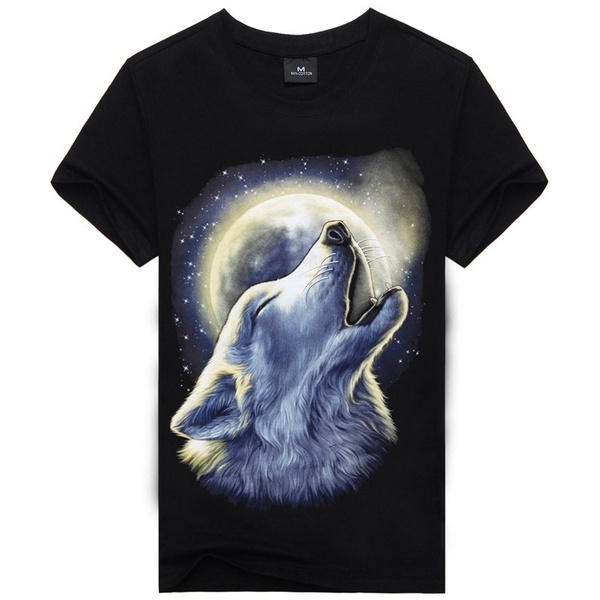}
\end{minipage}
&
\begin{minipage}{.3\linewidth}
    \includegraphics[height=10mm]{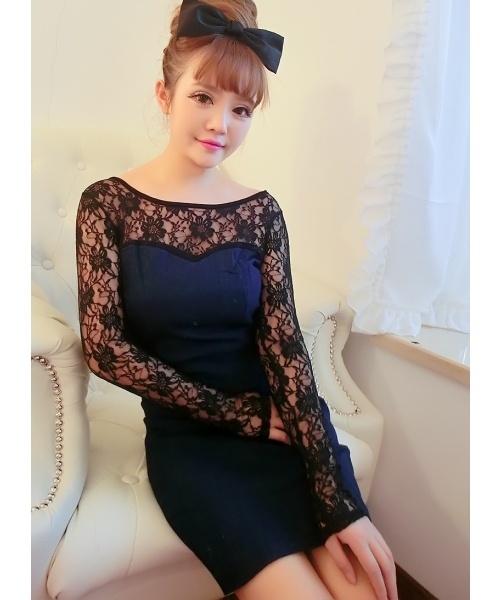}
\end{minipage} \\
Agent & \multicolumn{4}{L{0.8\linewidth}}{Does the property "Round Neck" apply to the product you want?} \\
User & \multicolumn{4}{L{0.8\linewidth}}{No} \\
Agent & \multicolumn{4}{L{0.8\linewidth}}{Here are some examples of the property "Lace"} \\
&
\begin{minipage}{.3\linewidth}
    \includegraphics[height=10mm]{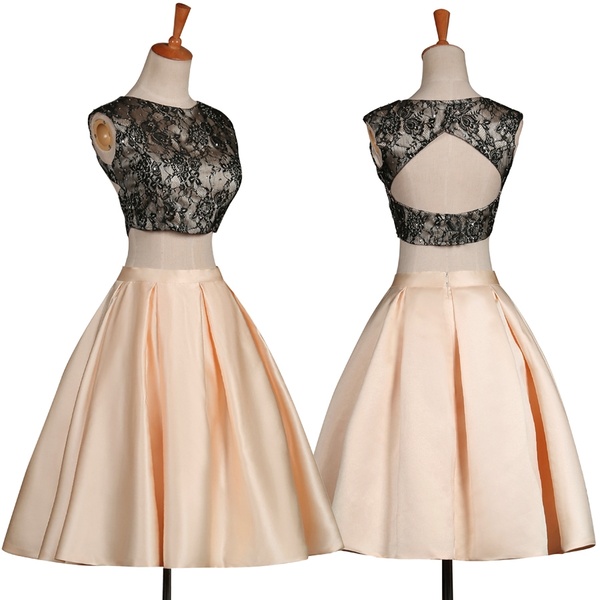}
\end{minipage}
&
\begin{minipage}{.3\linewidth}
    \includegraphics[height=10mm]{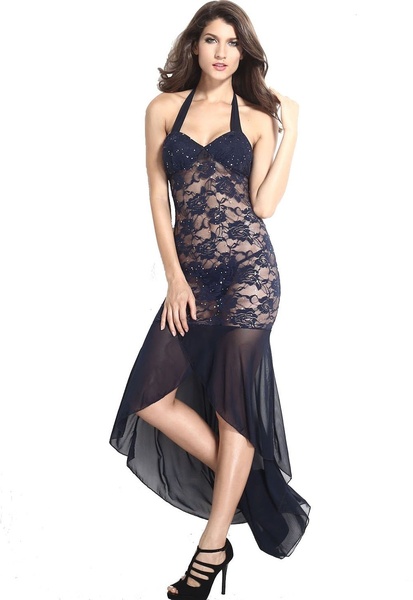}
\end{minipage}
&
\begin{minipage}{.3\linewidth}
    \includegraphics[height=10mm]{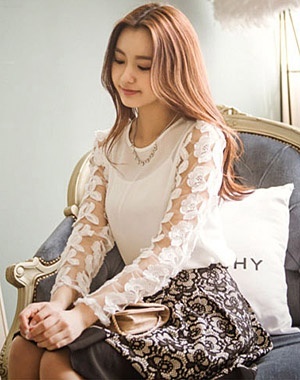}
\end{minipage} \\
Agent & \multicolumn{4}{L{0.8\linewidth}}{Does the property "Lace" apply to the product you want?} \\
User & \multicolumn{4}{L{0.8\linewidth}}{No} \\
Agent & \multicolumn{4}{L{0.8\linewidth}}{Here are some examples of the property "Teal"} \\
&
\begin{minipage}{.3\linewidth}
    \includegraphics[height=10mm]{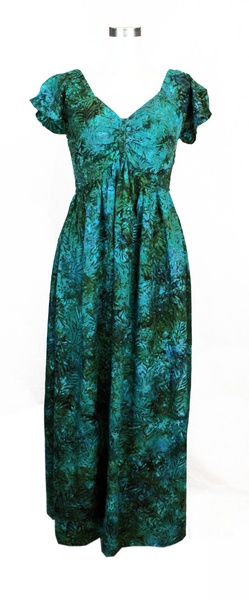}
\end{minipage}
&
\begin{minipage}{.3\linewidth}
    \includegraphics[height=10mm]{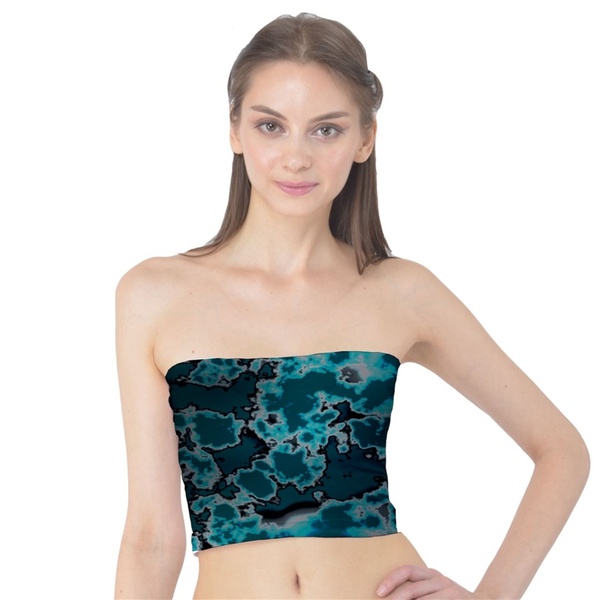}
\end{minipage}
&
\begin{minipage}{.3\linewidth}
    \includegraphics[height=10mm]{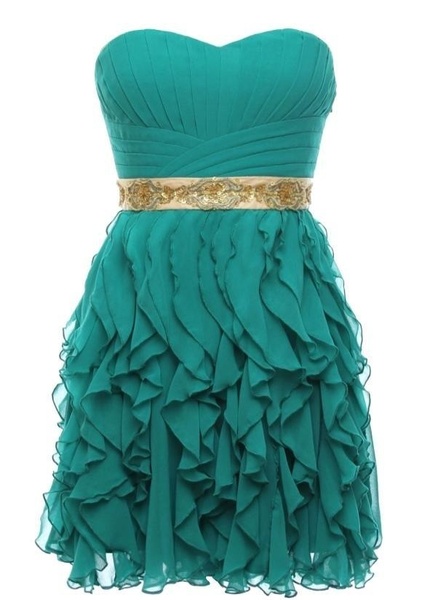}
\end{minipage} \\
Agent & \multicolumn{4}{L{0.8\linewidth}}{Does the property "Teal" apply to the product you want?} \\
User & \multicolumn{4}{L{0.8\linewidth}}{No} \\
Agent & \multicolumn{4}{L{0.8\linewidth}}{Here are some examples of the property "Paisley"} \\
&
\begin{minipage}{.3\linewidth}
    \includegraphics[height=10mm]{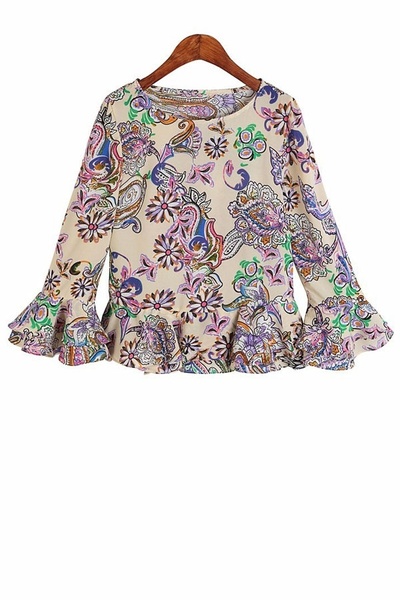}
\end{minipage}
&
\begin{minipage}{.3\linewidth}
    \includegraphics[height=10mm]{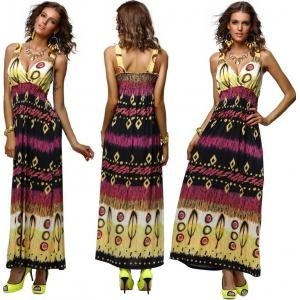}
\end{minipage}
&
\begin{minipage}{.3\linewidth}
    \includegraphics[height=10mm]{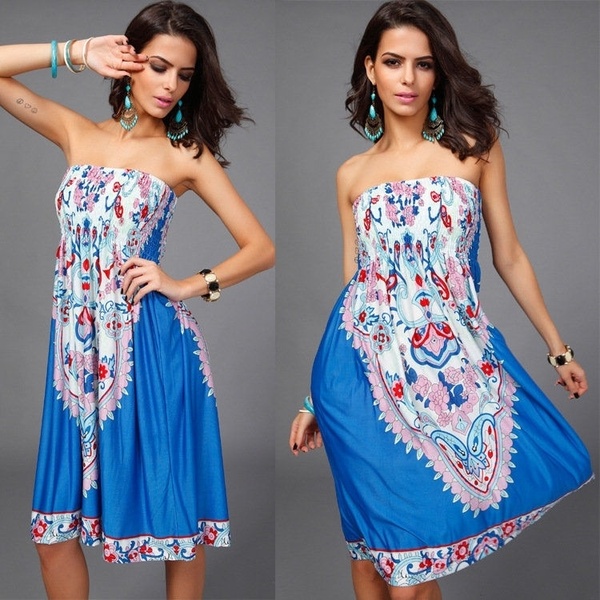}
\end{minipage} \\
Agent & \multicolumn{4}{L{0.8\linewidth}}{Does the property "Paisley" apply to the product you want?} \\
User & \multicolumn{4}{L{0.8\linewidth}}{No} \\
Agent & \multicolumn{4}{L{0.8\linewidth}}{Here are some examples of the property "Long Sleeved"} \\
&
\begin{minipage}{.3\linewidth}
    \includegraphics[height=10mm]{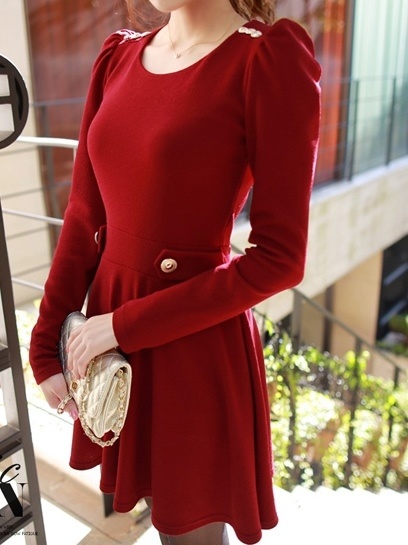}
\end{minipage}
&
\begin{minipage}{.3\linewidth}
    \includegraphics[height=10mm]{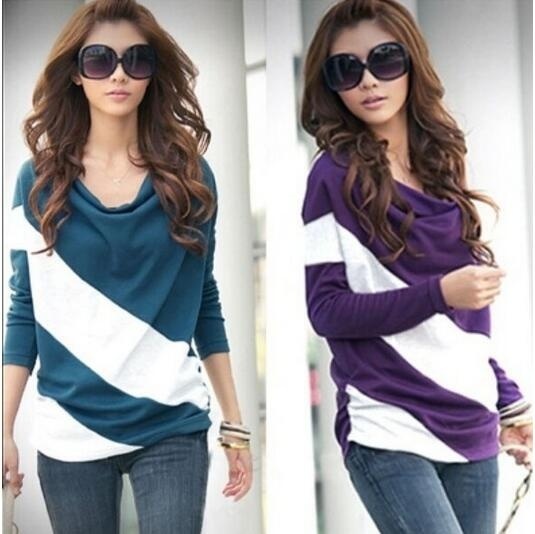}
\end{minipage}
&
\begin{minipage}{.3\linewidth}
    \includegraphics[height=10mm]{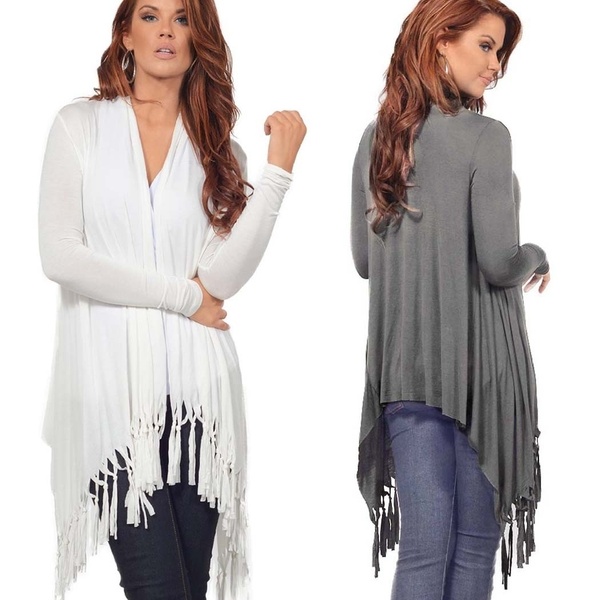}
\end{minipage} \\
Agent & \multicolumn{4}{L{0.8\linewidth}}{Does the property "Long Sleeved" apply to the product you want?} \\
User & \multicolumn{4}{L{0.8\linewidth}}{No} \\
Agent & \multicolumn{4}{L{0.8\linewidth}}{Here are some examples of the property "Quilted"} \\
&
\begin{minipage}{.3\linewidth}
    \includegraphics[height=10mm]{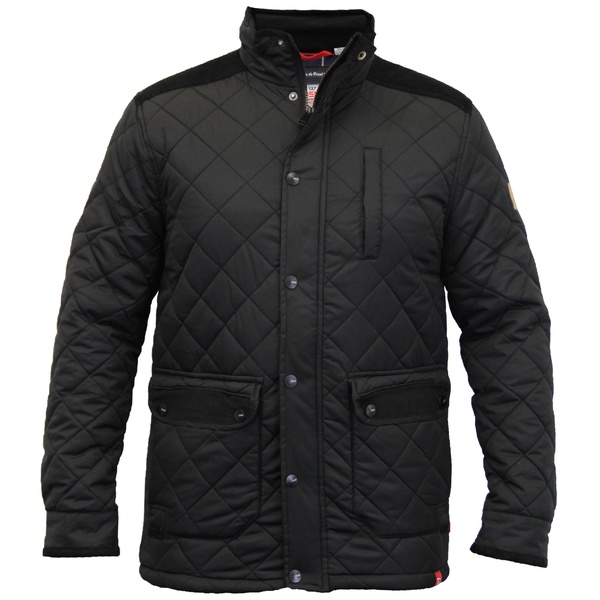}
\end{minipage}
&
\begin{minipage}{.3\linewidth}
    \includegraphics[height=10mm]{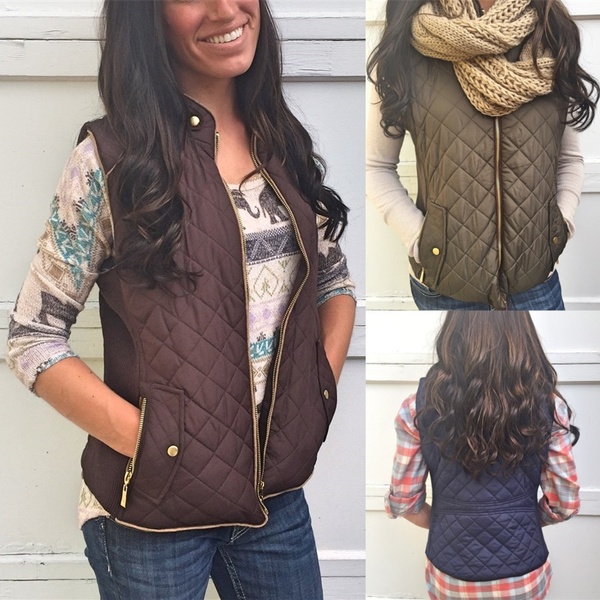}
\end{minipage}
&
\begin{minipage}{.3\linewidth}
    \includegraphics[height=10mm]{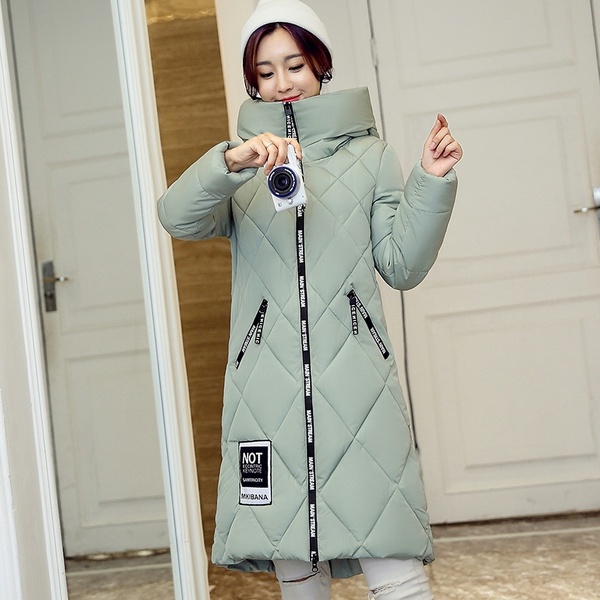}
\end{minipage} \\
Agent & \multicolumn{4}{L{0.8\linewidth}}{Does the property "Quilted" apply to the product you want?} \\
User & \multicolumn{4}{L{0.8\linewidth}}{No} \\
Agent & \multicolumn{4}{L{0.8\linewidth}}{Here are some examples of the property "Spaghetti Straps"} \\
&
\begin{minipage}{.3\linewidth}
    \includegraphics[height=10mm]{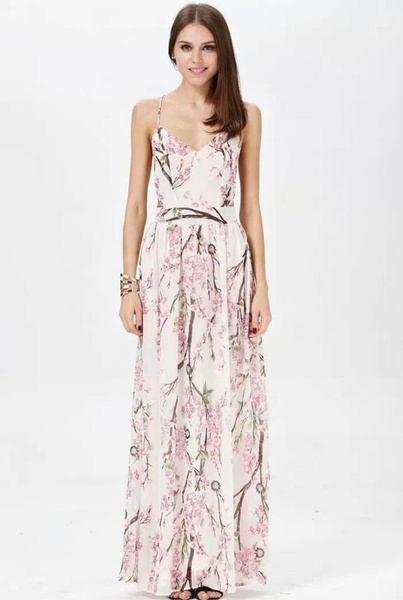}
\end{minipage}
&
\begin{minipage}{.3\linewidth}
    \includegraphics[height=10mm]{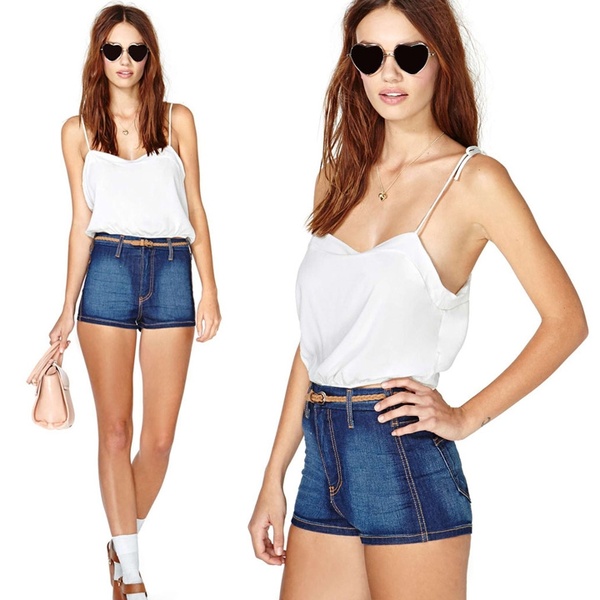}
\end{minipage}
&
\begin{minipage}{.3\linewidth}
    \includegraphics[height=10mm]{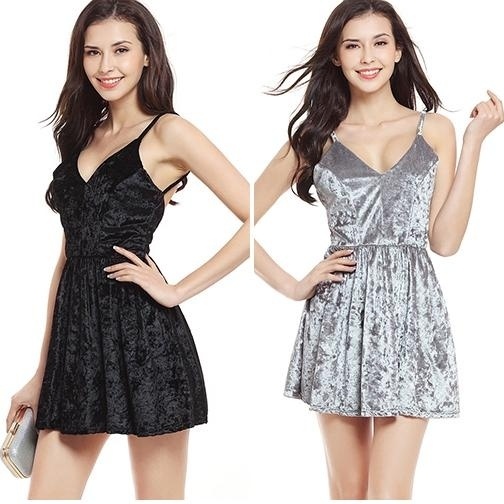}
\end{minipage} \\
Agent & \multicolumn{4}{L{0.8\linewidth}}{Does the property "Spaghetti Straps" apply to the product you want?} \\
User & \multicolumn{4}{L{0.8\linewidth}}{No} \\
Agent & \multicolumn{4}{L{0.8\linewidth}}{Here are some examples of the property "Tunic"} \\
&
\begin{minipage}{.3\linewidth}
    \includegraphics[height=10mm]{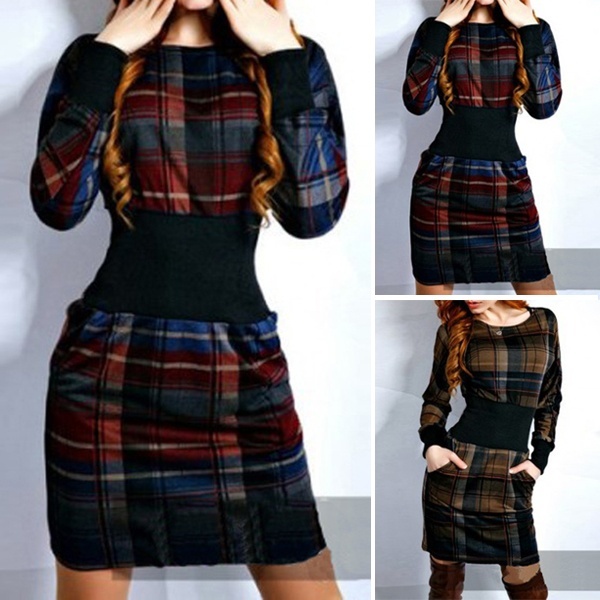}
\end{minipage}
&
\begin{minipage}{.3\linewidth}
    \includegraphics[height=10mm]{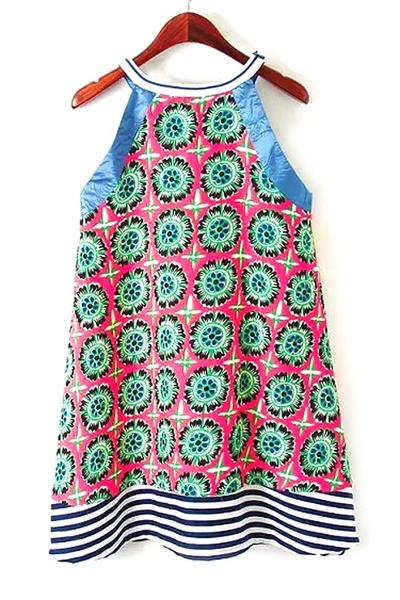}
\end{minipage}
&
\begin{minipage}{.3\linewidth}
    \includegraphics[height=10mm]{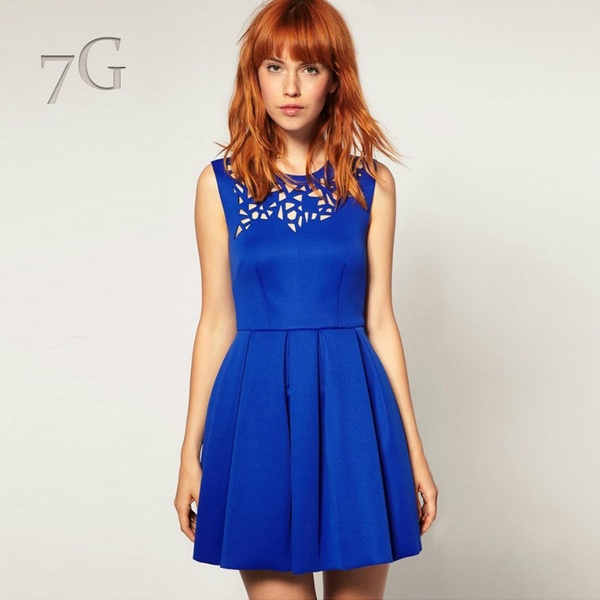}
\end{minipage} \\
Agent & \multicolumn{4}{L{0.8\linewidth}}{Does the property "Tunic" apply to the product you want?} \\
User & \multicolumn{4}{L{0.8\linewidth}}{No} \\
Agent & \multicolumn{4}{L{0.8\linewidth}}{Here are some examples of the property "Pin Stripes"} \\
&
\begin{minipage}{.3\linewidth}
    \includegraphics[height=10mm]{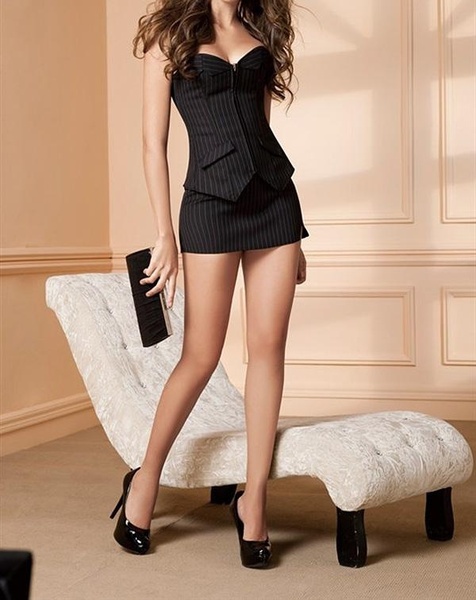}
\end{minipage}
&
\begin{minipage}{.3\linewidth}
    \includegraphics[height=10mm]{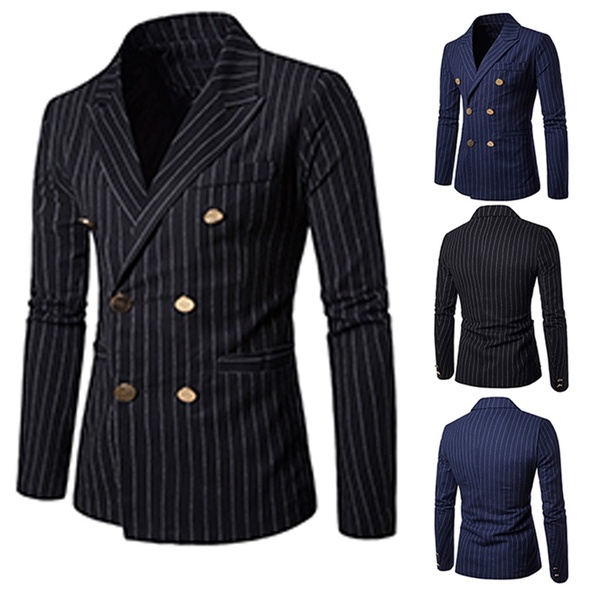}
\end{minipage}
&
\begin{minipage}{.3\linewidth}
    \includegraphics[height=10mm]{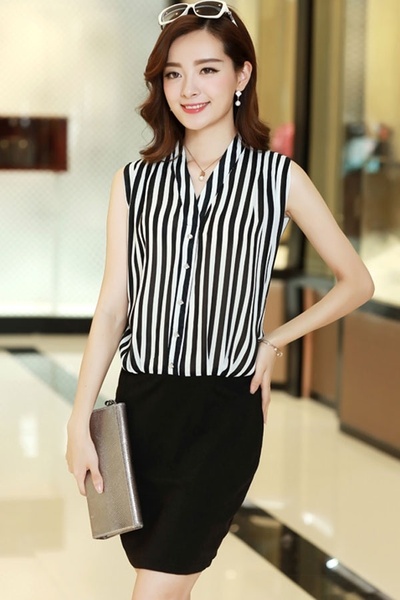}
\end{minipage} \\
Agent & \multicolumn{4}{L{0.8\linewidth}}{Does the property "Pin Stripes" apply to the image below?} \\
&
\begin{minipage}{.3\linewidth}
    \includegraphics[height=10mm]{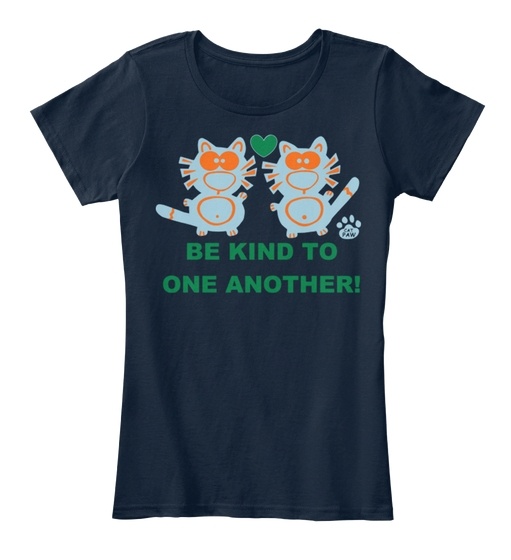}
\end{minipage} \\
User & \multicolumn{4}{L{0.8\linewidth}}{No} \\
Agent & \multicolumn{4}{L{0.8\linewidth}}{Here are some examples of the property "Pink"} \\
&
\begin{minipage}{.3\linewidth}
    \includegraphics[height=10mm]{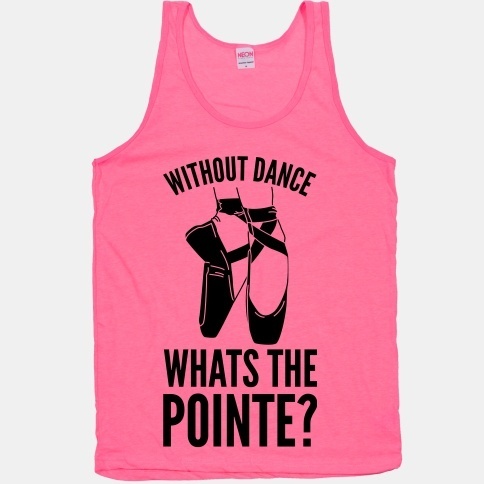}
\end{minipage}
&
\begin{minipage}{.3\linewidth}
    \includegraphics[height=10mm]{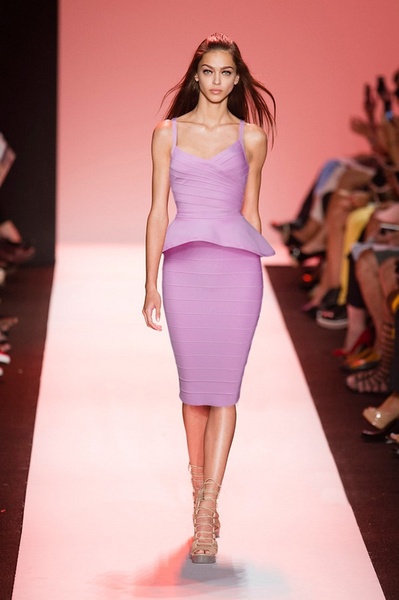}
\end{minipage}
&
\begin{minipage}{.3\linewidth}
    \includegraphics[height=10mm]{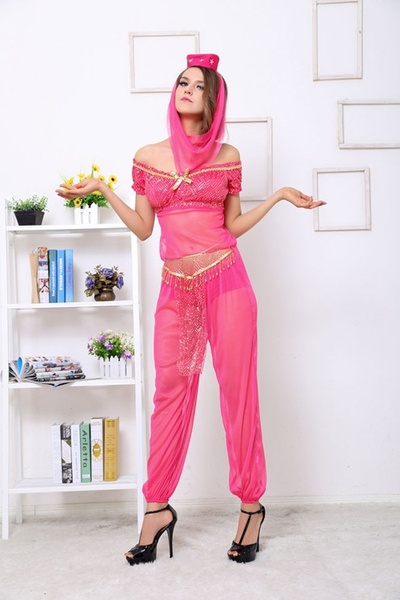}
\end{minipage} \\
Agent & \multicolumn{4}{L{0.8\linewidth}}{Does the property "Pink" apply to the image below?} \\
&
\begin{minipage}{.3\linewidth}
    \includegraphics[height=10mm]{figures/turk_dialog/276014.jpg}
\end{minipage} \\
User & \multicolumn{4}{L{0.8\linewidth}}{No} \\
Agent & \multicolumn{4}{L{0.8\linewidth}}{Here are some examples of the property "Organza"} \\
&
\begin{minipage}{.3\linewidth}
    \includegraphics[height=10mm]{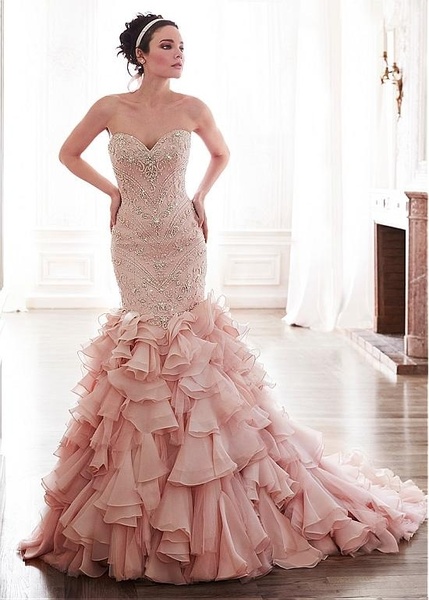}
\end{minipage}
&
\begin{minipage}{.3\linewidth}
    \includegraphics[height=10mm]{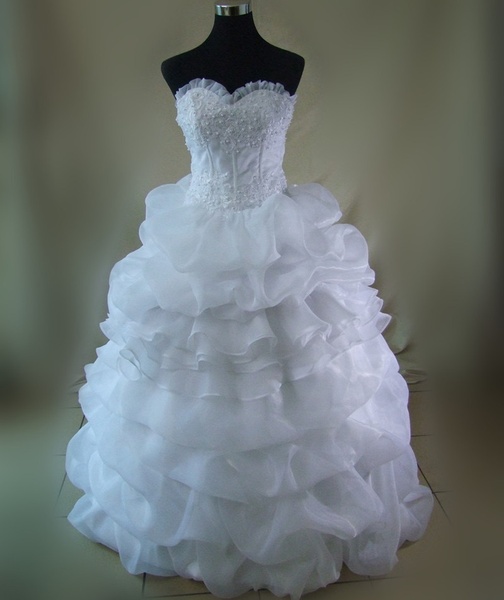}
\end{minipage}
&
\begin{minipage}{.3\linewidth}
    \includegraphics[height=10mm]{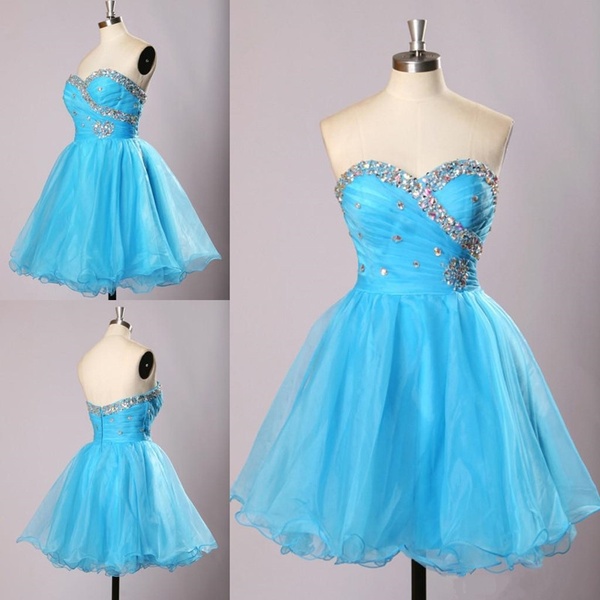}
\end{minipage} \\
Agent & \multicolumn{4}{L{0.8\linewidth}}{Does the property "Organza" apply to the image below?} \\
&
\begin{minipage}{.3\linewidth}
    \includegraphics[height=10mm]{figures/turk_dialog/276014.jpg}
\end{minipage} \\
User & \multicolumn{4}{L{0.8\linewidth}}{No} \\
Agent & \multicolumn{4}{L{0.8\linewidth}}{Here are some examples of the property "Heels"} \\
&
\begin{minipage}{.3\linewidth}
    \includegraphics[height=10mm]{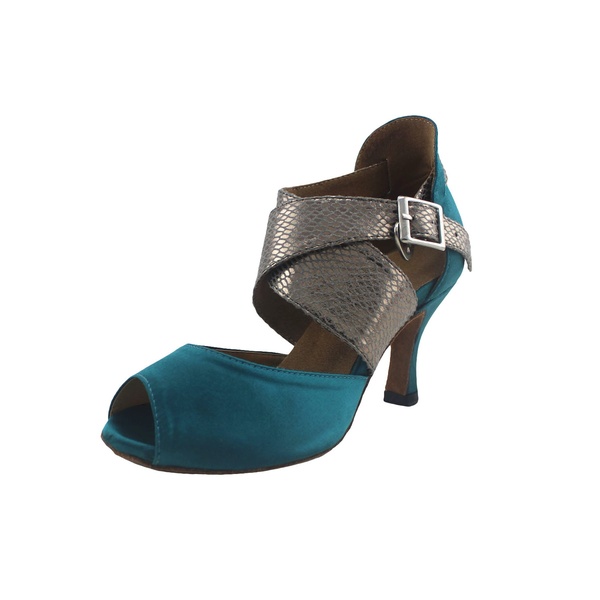}
\end{minipage}
&
\begin{minipage}{.3\linewidth}
    \includegraphics[height=10mm]{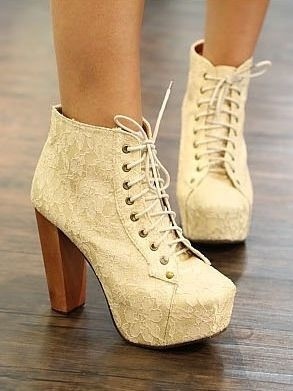}
\end{minipage}
&
\begin{minipage}{.3\linewidth}
    \includegraphics[height=10mm]{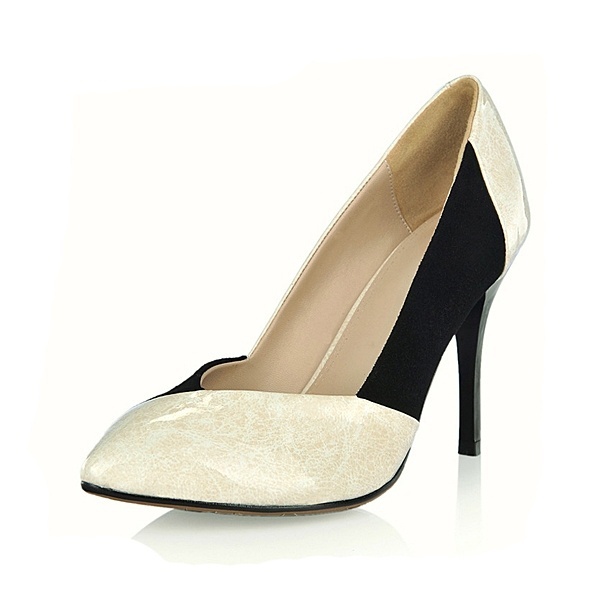}
\end{minipage} \\
Agent & \multicolumn{4}{L{0.8\linewidth}}{Does the property "Heels" apply to the image below?} \\
&
\begin{minipage}{.3\linewidth}
    \includegraphics[height=10mm]{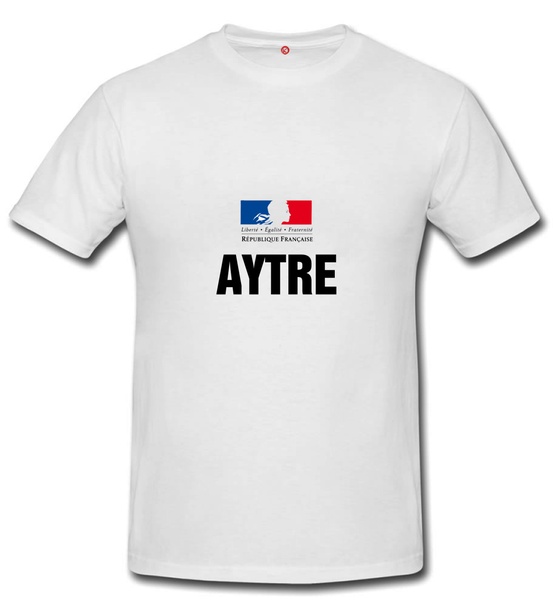}
\end{minipage} \\
User & \multicolumn{4}{L{0.8\linewidth}}{No} \\
Agent & \multicolumn{4}{L{0.8\linewidth}}{Here are some examples of the property "Sneakers"} \\
&
\begin{minipage}{.3\linewidth}
    \includegraphics[height=10mm]{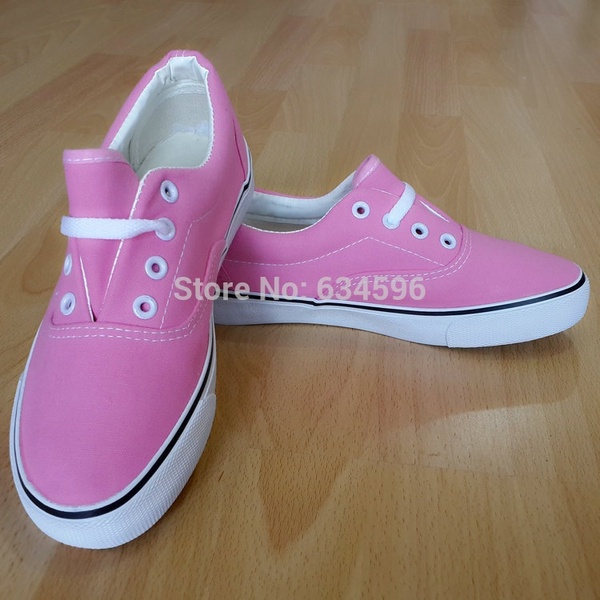}
\end{minipage}
&
\begin{minipage}{.3\linewidth}
    \includegraphics[height=10mm]{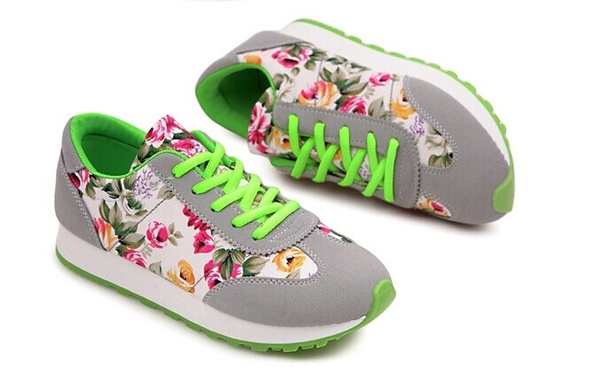}
\end{minipage}
&
\begin{minipage}{.3\linewidth}
    \includegraphics[height=10mm]{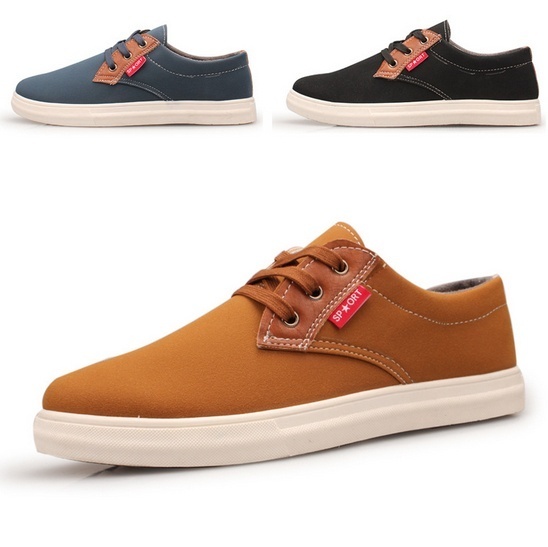}
\end{minipage} \\
Agent & \multicolumn{4}{L{0.8\linewidth}}{Does the property "Sneakers" apply to the image below?} \\
&
\begin{minipage}{.3\linewidth}
    \includegraphics[height=10mm]{figures/turk_dialog/173079.jpg}
\end{minipage} \\
User & \multicolumn{4}{L{0.8\linewidth}}{No} \\
Agent & \multicolumn{4}{L{0.8\linewidth}}{Here are some examples of the property "Galaxy"} \\
&
\begin{minipage}{.3\linewidth}
    \includegraphics[height=10mm]{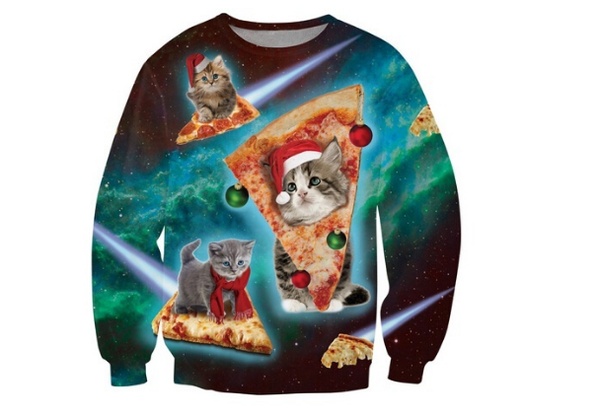}
\end{minipage}
&
\begin{minipage}{.3\linewidth}
    \includegraphics[height=10mm]{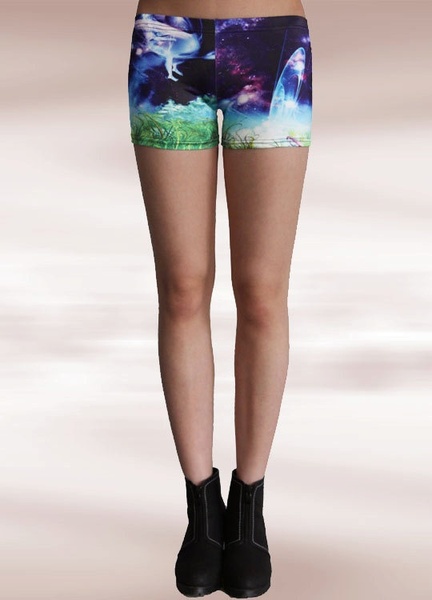}
\end{minipage}
&
\begin{minipage}{.3\linewidth}
    \includegraphics[height=10mm]{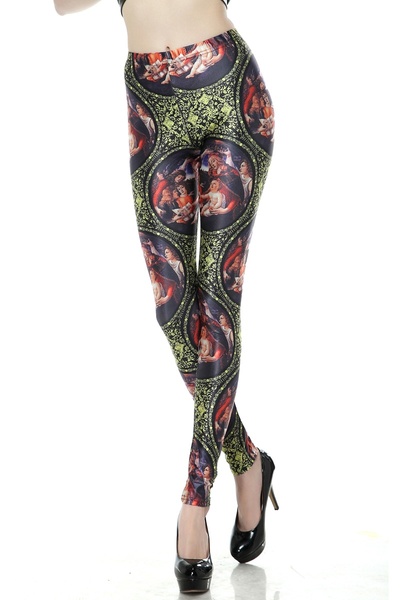}
\end{minipage} \\
Agent & \multicolumn{4}{L{0.8\linewidth}}{Does the property "Galaxy" apply to the image below?} \\
&
\begin{minipage}{.3\linewidth}
    \includegraphics[height=10mm]{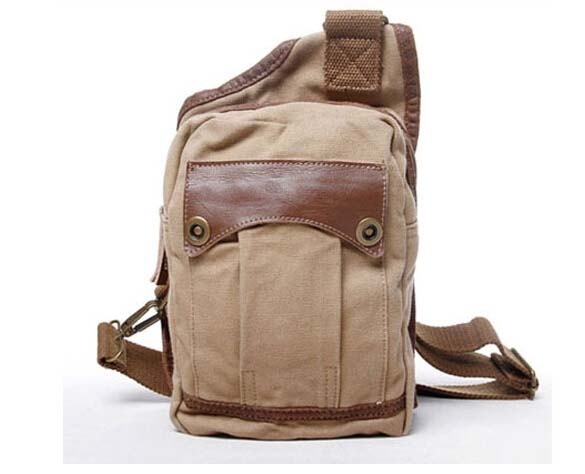}
\end{minipage} \\
User & \multicolumn{4}{L{0.8\linewidth}}{No} \\
Agent & \multicolumn{4}{L{0.8\linewidth}}{Here are some examples of the property "Checkered"} \\
&
\begin{minipage}{.3\linewidth}
    \includegraphics[height=10mm]{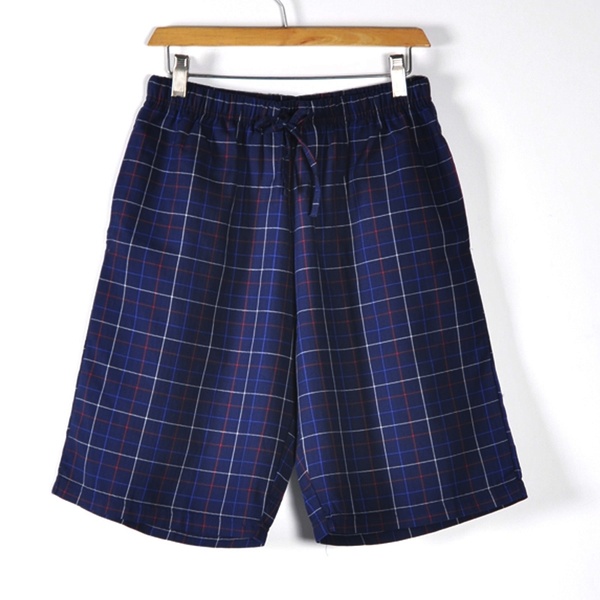}
\end{minipage}
&
\begin{minipage}{.3\linewidth}
    \includegraphics[height=10mm]{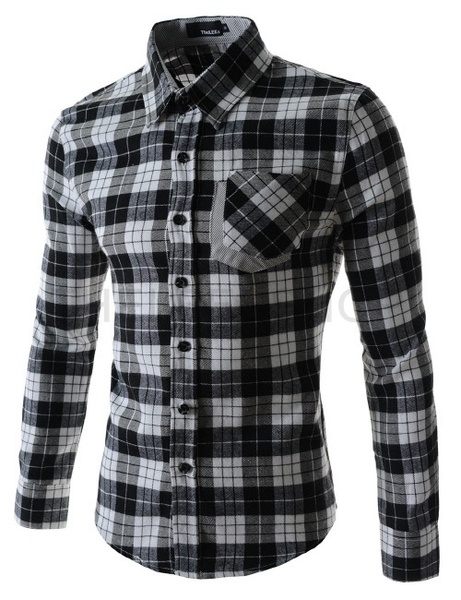}
\end{minipage}
&
\begin{minipage}{.3\linewidth}
    \includegraphics[height=10mm]{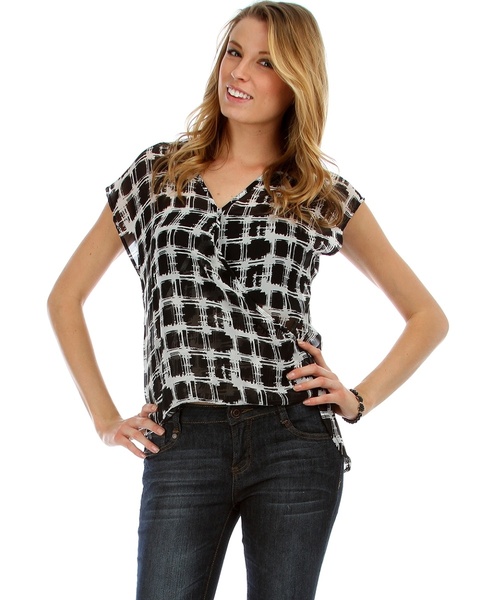}
\end{minipage} \\
Agent & \multicolumn{4}{L{0.8\linewidth}}{Does the property "Checkered" apply to the product you want?} \\
User & \multicolumn{4}{L{0.8\linewidth}}{No} \\\\
Agent & \multicolumn{4}{L{0.8\linewidth}}{Here are some examples of the property "Printed"} \\
&
\begin{minipage}{.3\linewidth}
    \includegraphics[height=10mm]{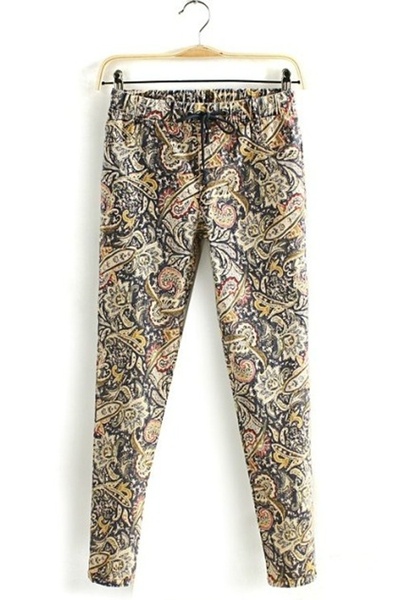}
\end{minipage}
&
\begin{minipage}{.3\linewidth}
    \includegraphics[height=10mm]{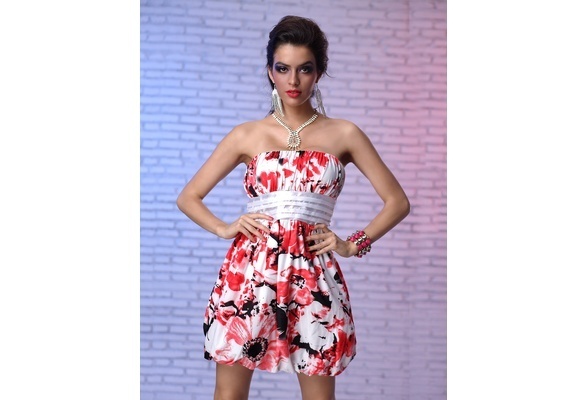}
\end{minipage}
&
\begin{minipage}{.3\linewidth}
    \includegraphics[height=10mm]{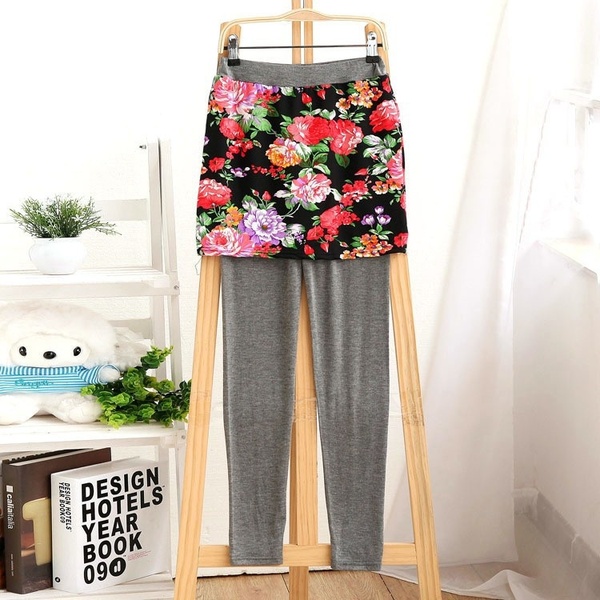}
\end{minipage} \\
Agent & \multicolumn{4}{L{0.8\linewidth}}{Does the property "Printed" apply to the product you want?} \\
User & \multicolumn{4}{L{0.8\linewidth}}{No} \\
Agent & \multicolumn{4}{L{0.8\linewidth}}{Here are some examples of the property "Shoe Accessories"} \\
&
\begin{minipage}{.3\linewidth}
    \includegraphics[height=10mm]{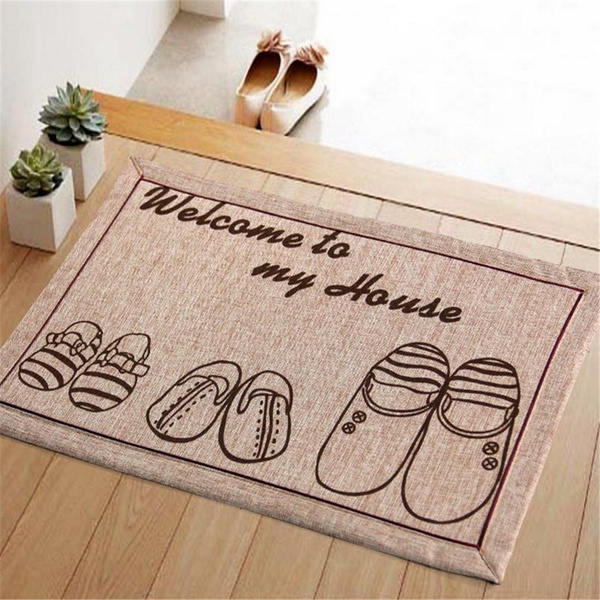}
\end{minipage}
&
\begin{minipage}{.3\linewidth}
    \includegraphics[height=10mm]{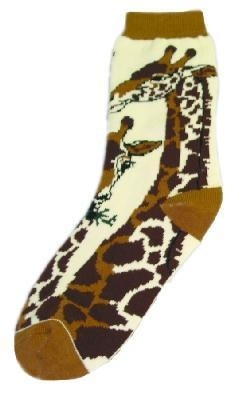}
\end{minipage}
&
\begin{minipage}{.3\linewidth}
    \includegraphics[height=10mm]{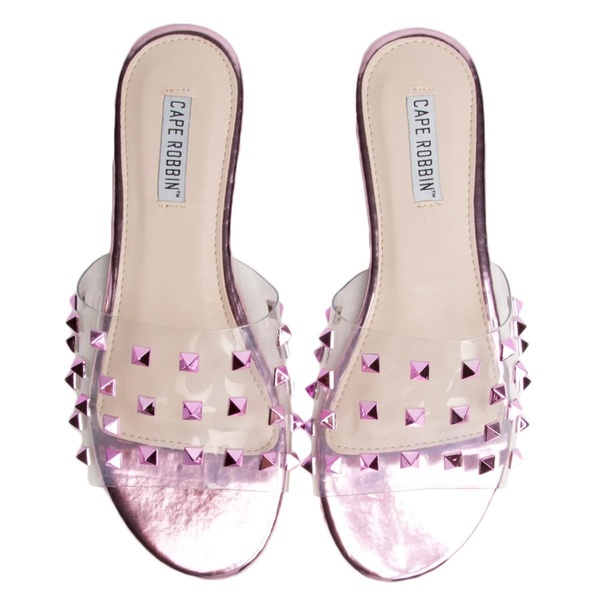}
\end{minipage} \\
Agent & \multicolumn{4}{L{0.8\linewidth}}{Does the property "Shoe Accessories" apply to the image below?} \\
&
\begin{minipage}{.3\linewidth}
    \includegraphics[height=10mm]{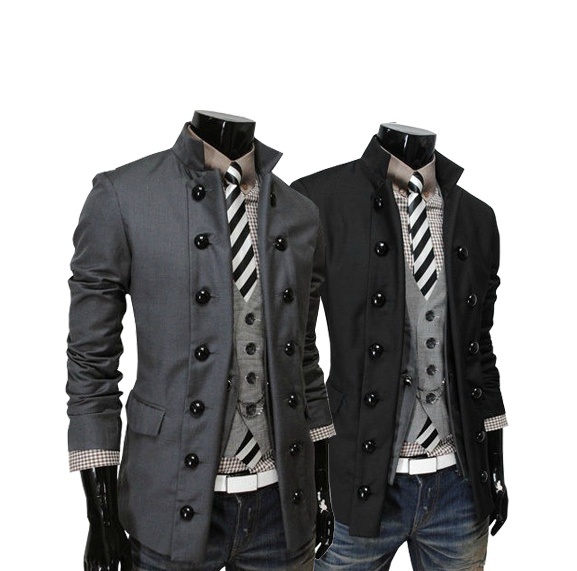}
\end{minipage} \\
User & \multicolumn{4}{L{0.8\linewidth}}{No} \\
Agent & \multicolumn{4}{L{0.8\linewidth}}{Here are some examples of the property "Satin"} \\
&
\begin{minipage}{.3\linewidth}
    \includegraphics[height=10mm]{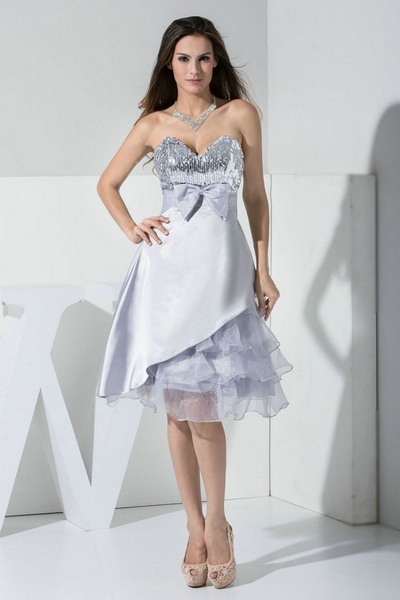}
\end{minipage}
&
\begin{minipage}{.3\linewidth}
    \includegraphics[height=10mm]{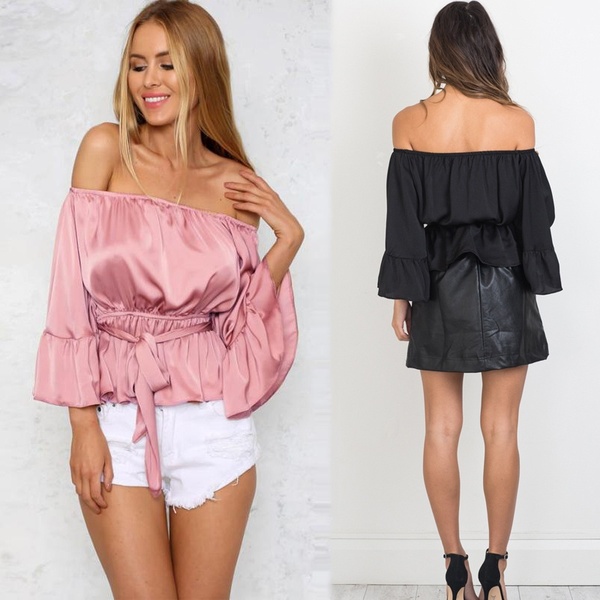}
\end{minipage}
&
\begin{minipage}{.3\linewidth}
    \includegraphics[height=10mm]{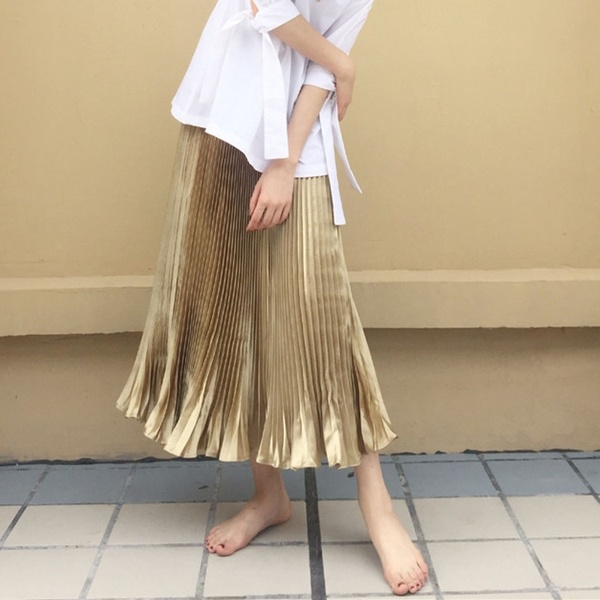}
\end{minipage} \\
Agent & \multicolumn{4}{L{0.8\linewidth}}{Does the property "Satin" apply to the image below?} \\
&
\begin{minipage}{.3\linewidth}
    \includegraphics[height=10mm]{figures/turk_dialog/718862.jpg}
\end{minipage} \\
User & \multicolumn{4}{L{0.8\linewidth}}{No} \\
Agent & \multicolumn{4}{L{0.8\linewidth}}{Is this the image you were looking for?} \\
&
\begin{minipage}{.3\linewidth}
    \includegraphics[height=10mm]{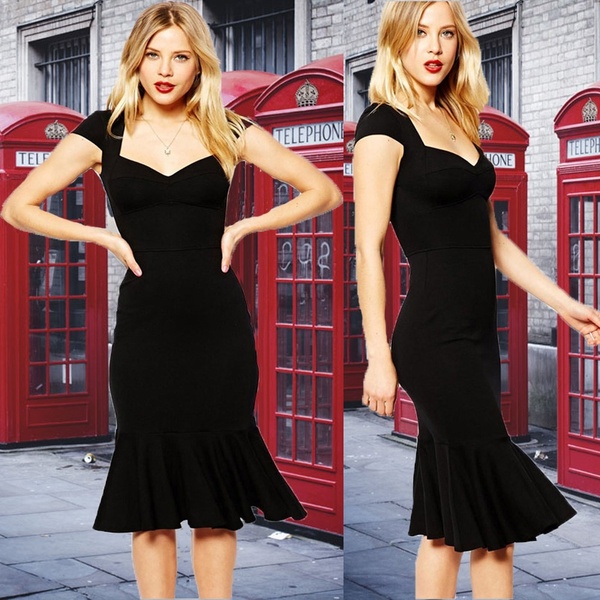}
\end{minipage} \\
User & \multicolumn{4}{L{0.8\linewidth}}{Yes}
\end{xtabular}

\end{document}